\documentclass[lettersize,journal]{IEEEtran}
\usepackage{amsmath,amsfonts}
\usepackage{algorithmic}
\usepackage{hyperref}
\usepackage{algorithm}
\usepackage{array}
\usepackage[caption=false,font=normalsize,labelfont=sf,textfont=sf]{subfig}
\usepackage{textcomp}
\usepackage{stfloats}
\usepackage{url}
\usepackage{verbatim}
\usepackage{graphicx}
\usepackage{cite}
\usepackage{enumitem}
\usepackage{multirow}
\usepackage{diagbox}
\usepackage{xcolor}
\usepackage{booktabs}
\usepackage{caption}
\usepackage{rotating}
\newcolumntype{C}[1]{>{\centering\arraybackslash}p{#1}}

\begin{document}
	
	\title{A Language Model-Driven Semi-Supervised Ensemble Framework for Illicit Market Detection Across Deep/Dark Web and Social Platforms}
	
	\author{
		Navid Yazdanjue$^{1}$,
		Morteza Rakhshaninejad$^{2}$,
		Hossein Yazdanjouei$^{3}$,
		Mohammad Sadegh Khorshidi$^{1}$,\\
		Mikko S. Niemelä$^{4}$,
		Fang Chen$^{1}$,
		Amir H. Gandomi$^{1,5,\dagger}$%
		\thanks{$^{1}$Data Science Institute, University of Technology Sydney, Sydney, Australia. $^{2}$Department of Data Analysis
and Mathematical Modeling, Ghent University, Ghent, Belgium. $^{3}$Department of Computer Science, Khazar University, Baku, Azerbaijan. $^{4}$Cyber Intelligence House (CIH), Singapore. $^{5}$University Research and Innovation Center (EKIK), Óbuda University, 1034, Budapest, Hungary. $\dagger$Corresponding author: Amir H. Gandomi (email: amir.gandomi@uts.edu.au).}
		%\thanks{Manuscript received April 19, 2021; revised August 16, 2021.}
	}
	
	\markboth{}%
	{Shell \MakeLowercase{\textit{et al.}}: A Sample Article Using IEEEtran.cls for IEEE Journals}
	
	
	\maketitle
	
	\begin{abstract}
		Illegal marketplaces have increasingly shifted to concealed parts of the internet, including the deep and dark web, as well as platforms such as Telegram, Reddit, and Pastebin. These channels enable the anonymous trade of illicit goods including drugs, weapons, and stolen credentials. Detecting and categorizing such content remains challenging due to limited labeled data, the evolving nature of illicit language, and the structural heterogeneity of online sources. This paper presents a hierarchical classification framework that combines fine-tuned language models with a semi-supervised ensemble learning strategy to detect and classify illicit marketplace content across diverse platforms. We extract semantic representations using ModernBERT, a transformer model for long documents, fine-tuned on domain-specific data from deep and dark web pages, Telegram channels, Subreddits, and Pastebin pastes to capture specialized jargon and ambiguous linguistic patterns. In addition, we incorporate manually engineered features such as document structure, embedded patterns including Bitcoin addresses, emails, and IPs, and metadata, which complement language model embeddings. The classification pipeline operates in two stages. The first stage uses a semi-supervised ensemble of XGBoost, Random Forest, and SVM with entropy-based weighted voting to detect sales-related documents. The second stage further classifies these into drug, weapon, or credential sales. Experiments on three datasets, including our multi-source corpus, DUTA, and CoDA, show that our model outperforms several baselines, including BERT, ModernBERT, DarkBERT, ALBERT, Longformer, and BigBird. The model achieves an accuracy of 0.96489, an F1-score of 0.93467, and a TMCC of 0.95388, demonstrating strong generalization, robustness under limited supervision, and effectiveness in real-world illicit content detection.
	\end{abstract}
	
	\begin{IEEEkeywords}
		Language models, Transformers, Semi--supervised learning, Ensemble learning, Dark Web.
	\end{IEEEkeywords}
	
	\section{Introduction}
\begin{figure}[!t]
		\centering
		\includegraphics[width=1\linewidth]{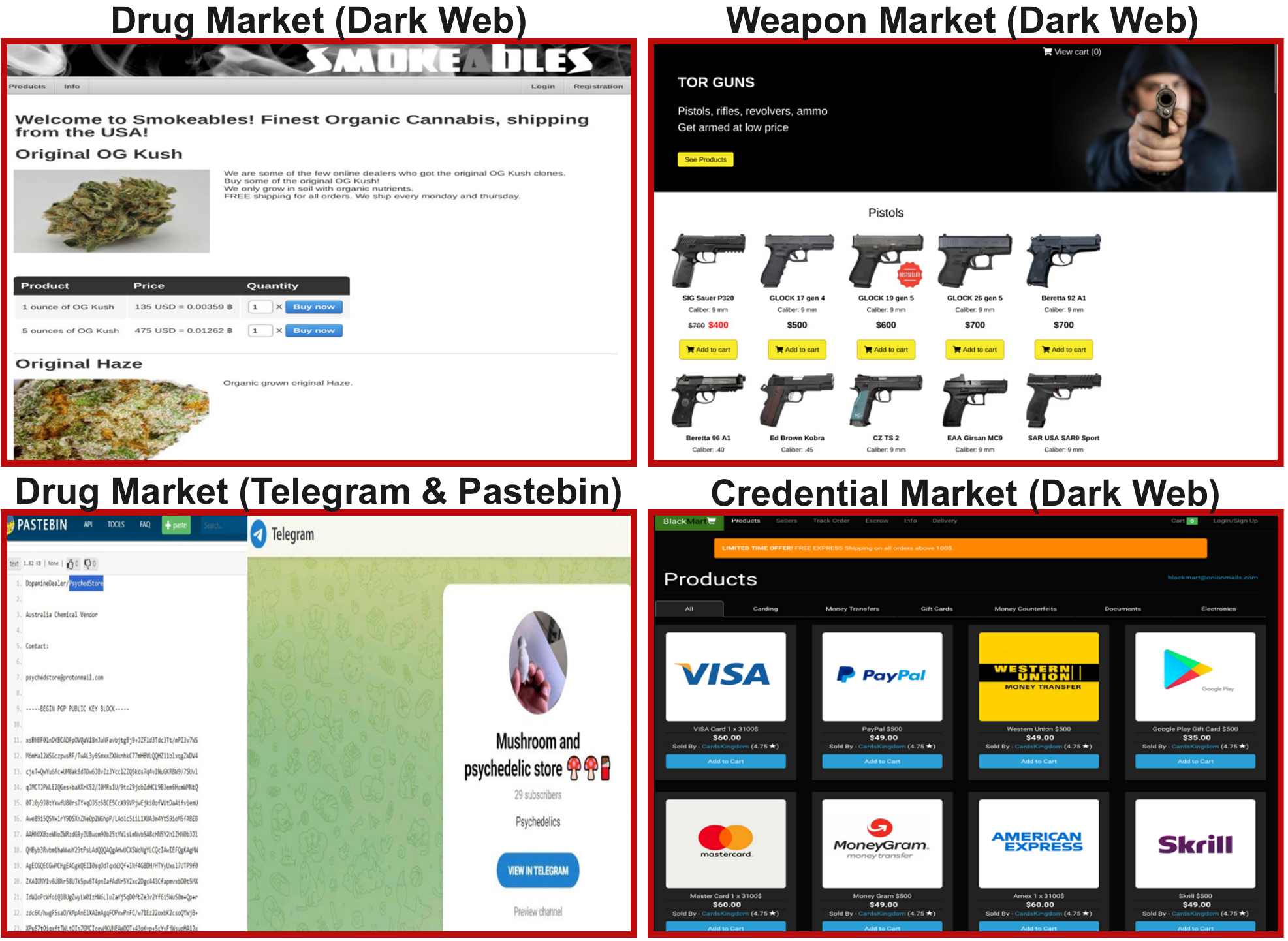}
		\caption{Examples of illicit marketplaces, such as drugs, weapons, and stolen credentials.}
		\label{fig:1}
	\end{figure}

	\IEEEPARstart{I}{n} today’s digital world, the expanse of the World Wide Web has introduced remarkable possibilities as well as formidable challenges. While much of the web remains openly accessible, commonly referred to as the surface web, a vast, less-visible portion known as the deep web also exists. The deep web includes areas of the internet that are not indexed by standard search engines (e.g., Google and Bing). While much of the deep web serves legitimate purposes, some areas are exploited for illegal activities. A prominent example is the dark web, which hosts hidden services that enable anonymous communication and trade, using layered encryption to mask user identities, locations, and transactions. These hidden services have given rise to a wide range of illegal marketplaces, where illicit goods such as drugs, weapons, and stolen credentials are openly advertised and traded \cite{haasio2020information}. While these marketplaces are often linked to the dark web, similar illicit activities have emerged on seemingly benign platforms like Telegram, Reddit, and content-hosting services such as Pastebin. Despite lacking built-in anonymity, these platforms are increasingly exploited for illegal marketplaces due to their accessibility, limited oversight, and rapid content dissemination. Fig.~\ref{fig:1} shows examples of illegal marketplaces on the hidden and surface web trading illicit products.
	
	The resilience of illicit marketplaces is evident in their continuous evolution and adaptability. Despite high-profile law enforcement interventions that have shut down major marketplaces, such as Silk Road (2013), AlphaBay (2017), DarkMarket (2021), Hydra Market (2022), and Genesis Market (2023) \cite{1-EUROPOL2023}, illegal trading has not only persisted but also grown. New platforms rapidly emerge, often emulating legitimate e-commerce interfaces to enhance usability while enabling the anonymous exchange of illicit goods. One such example is Rydox, a cybercrime marketplace seized by U.S. authorities most recently in December 2024, which facilitated the sale of stolen personal information and fraud tools \cite{2-Office2024}. The Rydox case highlights the rising complexity of illicit platforms and the continued efforts to dismantle them. Yet, new marketplaces keep emerging, using increasingly subtle tactics that make detection and classification a growing challenge for law enforcement and regulators.
	
	These challenges are intensified by the dynamic, fragmented, and unregulated nature of illicit marketplace content. Malicious actors often use ambiguous language, deliberate misspellings, and platform-specific jargon to evade detection, which significantly limits the effectiveness of conventional keyword-based monitoring tools. Additionally, the vast volume of unstructured textual data generated across these platforms presents a scalability challenge for manual or rule-based methods. Together, these characteristics highlight the urgent need for intelligent, automated detection systems capable of processing large-scale, heterogeneous textual content and extracting precise linguistic patterns that indicate illicit trading activities. To meet this need, many recent efforts have focused on supervised machine learning (ML) models, including deep learning architectures, which have shown promising results in identifying illicit content by learning complex linguistic patterns from labeled examples. However, their success heavily depends on the availability of large, high-quality labeled datasets. Such resources are particularly scarce in the context of deep/dark web and other mentioned platforms content, where manual annotation is time-consuming, costly, and requires expert knowledge. To overcome these limitations, unsupervised learning techniques such as clustering and topic modeling \cite{3-nazah2021unsupervised,4-porter2018analyzing,5-vahedi2021identifying} have been explored as alternatives that eliminate the need for labeled data. While these approaches can reveal hidden patterns and emerging trends within large text corpora, they often struggle with low classification accuracy and face challenges in accurately differentiating between illicit and non-illicit content.
	
	In addition to the scarcity of labeled data, a major limitation of existing approaches lies in the effectiveness of text representation techniques for modelling illicit marketplace content. Traditional natural language processing (NLP) techniques, such as TF-IDF \cite{6-salton1975vector}, Doc2Vec \cite{7-le2014distributed}, and GloVe \cite{8-pennington2014glove}, primarily rely on word frequency or co-occurrence patterns but fail to capture nuanced semantics, contextual dependencies, and evolving language patterns often employed in illicit communications. This is especially problematic when criminals manipulate language through obfuscation, slang, and jargon to avoid detection \cite{hou2022jargon}. Since these traditional methods treat words in isolation and ignore context, they often fall short in distinguishing legal from illegal content in noisy, real-world data.
	
	In contrast, modern language models (LMs) built on transformer architectures, have overcome many limitations of traditional techniques by capturing contextual dependencies and nuanced semantics. By analyzing meanings of words in relation to their surrounding context, they can better handle the obfuscated and context-dependent language often seen in illicit marketplaces. However, many widely used pre-trained LMs (PLM), such as BERT \cite{9-devlin2019bert}, ALBERT \cite{10-lan2019albert}, and RoBERTa \cite{11-liu2019roberta}, are trained on clean and general-domain surface web corpora (e.g., Wikipedia and BookCorpus). This introduces a domain mismatch, as the linguistic characteristics of content in deep/dark web forums, Telegram, Reddit, and Pastebin differ from those found in general-purpose training data. Therefore, these models often struggle to generalize effectively to the noisy and coded language found in illicit marketplaces. Another practical limitation of standard LMs lies in their input length constraints. Models, such as BERT, ALBERT, and RoBERTa, are restricted to processing sequences of up to 512 tokens, which is insufficient for lengthy posts found on dark web forums or subreddits. This truncation can lead to the loss of critical contextual patterns, such as embedded identifiers, pricing structures, or transaction instructions, ultimately reducing the detection accuracy of illicit activity in real-world applications.
	
	In response to these challenges, this study introduces a novel sequential classification framework that combines fine-tuned transformer-based LMs with semi-supervised and ensemble learning strategies to more effectively detect and categorize illicit marketplace content. Our approach is designed to first identify sales-related documents and further classify them into critical categories, including drug, weapon, and stolen credential sales. We implement this framework by extracting features using ModernBERT \cite{12-warner2024smarter}, a modern long-context PLM capable of processing sequences up to 8192 tokens, significantly surpassing the 512-token limit of standard models, such as BERT. This extended capacity enables the model to capture complete contextual patterns in lengthy posts typical of dark web forums and subreddits. Furthermore, we fine-tune ModernBERT on domain-specific data collected from the deep and dark web, Telegram, Reddit, and Pastebin, allowing it to adapt to the unique linguistic structures, obfuscated terms, and specialized jargon prevalent in illicit marketplace content. This domain adaptation produces context-rich embeddings that better capture the nuanced language and operational characteristics of real-world illegal marketplaces. Alongside LM-based embedding, we incorporate a complementary set of manually engineered features that capture documents’ structural patterns and specific items often overlooked by LMs. These include document formatting and structure (e.g., indentations and line width), embedded items (e.g., images, emails, IPs, and Bitcoin addresses), as well as metadata related to the document’s source and time. Derived from our previous study \cite{yazdanjue2025cyber}, these features are particularly valuable for detecting and categorizing the illicit marketplace content, where key indicators may appear in non-linguistic, irregular, or visually structured formats. By combining these manually extracted features with the contextual embeddings from fine-tuned ModernBERT, our framework gains a more comprehensive view of each document, enhancing the accuracy of illegal marketplaces detection and classification.
	
	Building on this multi-faceted representation, we feed the concatenated features into a two-stage sequential document classification framework. The first stage is based on a Semi-Supervised Ensemble (SSE) model to identify sales-related documents across the platforms under study. This model addresses the challenge of labeled data scarcity through a semi-supervised approach that integrates a small labeled set with a large amount of unlabeled data \cite{13-chapelle2006semi,14-zhu2009introduction}, significantly reducing reliance on manual annotation. Also, the proposed model’s ensemble architecture, comprising XGBoost (XGB), Random Forest (RF), and Support Vector Machine (SVM) base learners (BLs), improves its robustness and generalization capabilities by aggregating the strengths of different classifiers. To further enhance decision reliability, the SSE model incorporates a novel entropy-based weighted voting mechanism that dynamically adjusts the influence of each BL, leading to more accurate pseudo-labels and improved classification of sales-related content. In the second stage, documents identified as sales-related are passed to three specialized semi-supervised XGB classifiers, each dedicated to one of the target illicit trade categories (i.e., drug sales, weapon sales, and stolen credential sales). The proposed sequential framework facilitates fine-grained classification by breaking down the complex multi-label task into more focused sub-problems, enabling more accurate identification of specific illicit marketplace activities. The key contributions of this study are summarized as follows:
	\begin{itemize}
		\item A diverse, multi-source dataset is constructed by aggregating content from the deep/dark web, Telegram, Reddit, and Pastebin, enabling realistic and comprehensive evaluation across heterogeneous platforms.
		\item To extract meaningful features from this heterogeneous dataset, ModernBERT is fine-tuned on domain-specific data from illicit sources to produce rich embeddings that capture the obfuscated and jargon-heavy language used in illegal marketplaces.
		\item A set of manually engineered features is concatenated with the ModernBERT embeddings, that capture structural layout, embedded items, and metadata associated with each document, enriching the representation beyond what LMs capture.
		\item Building on this enriched feature representation, a sequential classification framework is proposed that first detects sales-related documents and then classifies them into drug, weapon, or credential sales, enabling structured and effective hierarchical classification of illicit marketplace content.
		\item In the first stage, an SSE model is employed, combining XGB, RF, and SVM models within a self-training paradigm to improve robustness and accuracy under low-resource, noisy conditions.
		\item To further enhance the SSE model’s effectiveness, a novel weighted voting strategy is introduced that adjusts the influence of each BL based on its prediction confidence. Specifically, two entropy-based criteria, Mean Entropy of Correct Predictions ($MEC$) and Mean Entropy of Wrong Predictions ($MEW$), are defined to dynamically weight classifier outputs. This mechanism improves the reliability of pseudo-labels and increases the overall accuracy of sales-related content detection.
		\item In the second stage, three distinct self-training semi-supervised models are implemented, each based on an XGB classifier, to further categorize sales-related documents into specific illicit trade types, including drug, weapon, and credential sales.
	\end{itemize}
	
	The remainder of this paper is organized as follows. Section \ref{sec:litrev} provides an overview of related work in detecting and classifying illicit marketplace content using machine learning and natural language processing techniques. In Section \ref{sec:methodology}, we present the methodology, including dataset, feature extraction, and the design of our sequential classification framework. Section \ref{sec:exset} discusses the experimental setup, evaluation metrics, and results, highlighting the effectiveness of our approach. Finally, Section \ref{sec:conclusion} concludes the paper with key findings and potential directions for future research.
	
	\section{LITERATURE REVIEW}\label{sec:litrev}
	The increasing presence of illicit activities on the deep/dark web, as well as on surface platforms, such as Telegram, Reddit, and Pastebin, has intensified the demand for automated, intelligent methods to detect and categorize such content. In response, a growing body of research has explored ML approaches to extract and classify cyber threat intelligence from unstructured web data, which is outlined in Table~\ref{tab:1}. This table categorizes prior work based on the type of text representation used, including traditional statistical representations, static embeddings, and contextual transformer-based embeddings. It provides a comparative overview of each study’s data sources, feature extraction techniques, classification models, target categories, and key findings.
	\begin{table*}[!t]
		\centering
		\caption{Summary of Illicit Content Classification Studies across Deep/Dark Web and Related Platforms.}
		\label{tab:1}
		{\fontsize{7pt}{7pt}\selectfont % custom font size
			\renewcommand{\arraystretch}{1.3}
			\setlength{\tabcolsep}{1pt}
			\begin{tabular}{|C{1.6cm}|C{0.6cm}|C{1.5cm}|C{3.2cm}|C{2.3cm}|C{3.5cm}|C{5cm}|}
				\hline
				\textbf{Categories} &
				\textbf{Study} &
				\textbf{Data Source} &
				\textbf{Feature Extraction Technique} &
				\textbf{Classification Model} &
				\textbf{Target Categories} &
				\textbf{Key Results} \\ \hline
				
				\multirow{8}{=}{\centering Traditional\\ Statistical\\ Representations} 		
				& \cite{16-nunes2016darknet} & Deep/Dark Web Forums, Marketplaces & TF-IDF & NB, RF, SVM, LR & Market Products (Hacking, Drugs, Porno, Weapons, Software Service), Forum Discussions (Programming, Hacking, Cyber-security) & SVM with a linear kernel performed best, achieved 87\% recall and 85\% precision. \\ \cline{2-7} 
				& \cite{17-al2017classifying} & Dark Web Webpages & TF-IDF, BoW & LR, SVM, NB & 9 categories, such as Drugs, Child Porno, Counterfeit Credit-Cards, Cryptography, Hacking, Violence, Others & They introduced DUTA dataset, and TF-IDF with LR achieved the highest performance on categorizing the illicit contents. \\ \cline{2-7} 
				& \cite{18-kobayashi2020expert} & Dark Web Forums & BoW & Traditional Rule-based Technique & Credit card, Counterfeit, Drug, Hacking, Porno, Other  & Their system achieved 72.5\% recall for classification, limited by keyword reliance causing misclassification in ambiguous posts. \\ \cline{2-7} 
				& \cite{3-nazah2021unsupervised} & Dark Web Forums & TF-IDF & Unsupervised K-means Clustering with RF, SVM, NB, LR & Breach, Financial, Drug, Vendor, Account, Other, Product & Among the Models, K-means Clustering with RF achieved the best performance overall, followed closely by SVM, while LR showed weaker performance. \\ \cline{2-7} 
				& \cite{19-murty2022building} & Dark Web Webpages & TF-IDF & RF, LR, NB, SVM & Adult, Counterfeits, Market, Cryptocurrency, Drug, Services, Weapons & The best model was LR, with an Accuracy of 94\%, followed by RF and SVM, both scoring 91\% \\ \cline{2-7} 
				& \cite{20-dalvi2023hybrid} & Deep/Dark Web & TF-IDF & RNN, BR, CC, LP & Business, Cybersecurity, Education, Entertainment, Finance, Food, Health, Literature, Nature, Network, Politics, Security, Shopping & The RNN model outperformed BR, CC, LP, achieving the macro-Precision of 0.72. \\ \cline{2-7} 
				& \cite{21-qian2023prediction} & Dark Web Forums & TF-IDF & Unsupervised LDA Model with NB, LR, SVM, RF, KNN, MLP & Technology, Setting Pages,
				Node Selling, File Sharing, Porno, Search Pages, News, Unknown Pages
				& LDA model with the SVM classifier outperformed others with a 93\% Accuracy score. \\ \hline
				
				\multirow{4}{=}{\centering Static\\ Embeddings}
				& \cite{22-deliu2017extracting} & Dark Web Forums & Word2Vec, GloVe, Trigrams, BoW & SVM, DT, KNN, CNN & Credentials, Kyloggers, DDOS, Trojans, RATs, Spamming, Sql Injection, Crypters, Not-related & SVM with Trigrams achieved 98.62\% accuracy, slightly outperforming CNNs with Word2Vec and GloVe. \\ \cline{2-7} 
				& \cite{23-tavabi2018darkembed} & Dark Web Forums & DarkEmbed (Trained from scratch using Doc2Vec model on Deep/Dark web posts), TF-IDF & SVM, RF & Exploits vs. Not Exploits & DarkEmbed with an SVM outperformed traditional TF-IDF approaches with SVM or RF, achieving an F1 Score of 0.66 and AUC of 0.84. \\ \cline{2-7} 
				& \cite{24-kadoguchi2019exploring} & Dark Web Forums & Doc2Vec & MLP & Malware Offering & Doc2Vec + MLP achieved 93.9\% Accuracy in training but dropped to 79.4\% on unseen data. \\ \cline{2-7} 
				& \cite{25-ampel2021distilling} & Dark Web Forums & Hack2Vec (Static Embedding Distilled from BERT), Word2Vec, GloVe, fastText, BERT & BiLSTM & Web Application, DOS, Remote, Local, SQL Injection, Cross-site Scripting, File Inclusion, Overflow & Hack2Vec with BiLSTM classifier achieves the F1 Score of 66.52\%, outperformed GloVe, fastText, Word2Vec, and BERT with BiLSTM. \\ \hline
				
				\multirow{7}{=}{\centering Contextual\\ Transformer-based\\ Embeddings}
				& \cite{26-mendsaikhan2020identification} & Reddit & BERT (Fine-tuned), Doc2Vec (Trained from the Scratch) & Cosine similarity as a decision rule, LR & Cybersecurity-Relevant Posts & BERT achieved 91\% F1 Score, outperforming Doc2Vec with F1 Score of 0.63 \\ \cline{2-7} 
				& \cite{5-vahedi2021identifying} & Paste sites & BERT & Unsupervised LDA Model & Hackers, Malware, Network Exploits, Website Exploits, PII  & BERT-LDA outperformed standard LDA with significantly lower perplexity across paste sites. \\ \cline{2-7} 
				& \cite{27-jin2022shedding} & Dark Web Webpages & BERT, GloVe, TF-IDF & SVM, CNN & Porno, Drugs, Financial, Gambling, Cryptocurrency, Hacking, Arms/Weapons, Violence, Electronics, Others & They introduced CoDA dataset, and BERT achieved the highest performance (macro F1 Score of 92.49\%), followed by SVM with TF-IDF (91.19\%) and CNN with GloVe (87.23\%). \\ \cline{2-7} 
				& \cite{28-cascavilla2023illicit} & Dark Web Marketplaces & ULMFit, 
				Fine-tuned BERT, Fine-tuned RoBERTa, LSTM
				& The mentioned models were used as standalone classification models. & 19 categories, such as Drugs, Counterfeit, Credit-Cards, Fraud, Porno, Services, Cryptocurrency, Violence, Social Network & BERT achieved the highest macro-Accuracy of 96.08\% and 91.98\% for drug sub-categories, outperforming RoBERTa, ULMFit, and LSTM. \\ \cline{2-7} 
				& \cite{29-jin2023darkbert} & Dark Web Webpages & DarkBERT (A RoBERTa-based Model Pretrained from Scratch on Dark Web Pages), BERT, RoBERTa & The mentioned models were used as standalone classification models. & DUTA and CoDA Datasets’ Categories & DarkBERT, achieved the highest macro F1 Score (94.46\%) on the CoDA dataset, outperforming BERT and RoBERTa. \\ \cline{2-7} 
				& \cite{30-sangher2023towards} & Dark Web Marketplace & Fine-tuned BERT, GloVe & BERT, RNN, CNN, LSTM & Cybercrime, Not-Cybercrime, can’t Say & BERT model outperformed RNN, CNN, and LSTM models with GloVe in identifying cybercrime content, achieving the highest F1 Score of 0.79. \\ \cline{2-7} 
				& \cite{31-dalvi2024analysis} & Dark Web Webpages & BERT, Word2Vec, TF-IDF & BERTopic, Top2Vec, LDA, LSA, NMF & Hosting/Software, Hidden Wiki, Social Networks, Drugs, Business/Marketplace, Geography, Personal Health & BERTopic achieved the highest manual evaluation score (80\%) for topic coherence, accuracy, and non-redundancy, outperforming Top2Vec with Word2Vec, and LDA, LSA, and NMF with TF-IDF. \\ \hline
			\end{tabular}
		} % End tiny
		\captionsetup{justification=justified, singlelinecheck=false}
		\caption*{\footnotesize\textit{ Abbreviations used in the table include, NB = Naive Bayes; LR = Logistic Regression; RNN = Recurrent Neural Network; BR = Binary Relevance; CC = Classifier Chains; LP = Label Powerset; LDA = Latent Dirichlet Allocation; MLP = Multi-Layer Perceptron; DT = Decision Tree; CNN = Convolutional Neural Network; BiLSTM = Bidirectional Long Short-Term Memory; ULMFit = Universal Language Model Fine-tuning; BERT = Bidirectional Encoder Representations from Transformers; RoBERTa = Robustly Optimized BERT Approach; NMF = Non-negative Matrix Factorization; LSA = Latent Semantic Analysis.}}
	\end{table*}

	Despite notable progress in applying ML to detect illicit content on the dark web and related platforms, several key limitations remain in existing research. Most studies rely on narrow, single-platform datasets (e.g., forums, marketplaces, or Reddit), failing to capture the heterogeneity of real-world cybercrime sources. Also, many approaches still use traditional representations such as TF-IDF or static embeddings like Doc2Vec, which are inadequate for modeling the evolving, jargon-heavy language of illicit markets. Although transformer-based models like BERT and RoBERTa have been explored, they are rarely fine-tuned on dark web-specific data, limiting their contextual relevance. These models also face input length limits of 512 tokens, which hinders their ability to process long posts and results in the loss of important context. Additionally, structural layout, embedded items, and metadata, which have shown utility in detecting illicit activity, are also often overlooked in prior studies. Moreover, most prior work relies on fully supervised learning, which requires large labeled datasets that are costly and difficult to obtain, while semi-supervised learning remains underexplored, despite its promise in low-resource settings. In addition, most studies use single classifiers and do not employ ensemble methods that could combine diverse model strengths to improve robustness and adaptability.
    
	To overcome current limitations in illicit content detection, this study presents a two-stage classification framework that combines fine-tuned ModernBERT embeddings with manually engineered features. Also, a Semi-Supervised Ensemble model, integrating XGB, RF, and SVM with a novel weighted voting strategy, is used to detect sales-related content. The identified sales documents are then classified into key illicit categories (drug, weapon, and stolen credential sales) using specialized semi-supervised classifiers, enabling accurate identification across the diverse platforms under study.
     	
	\section{PROPOSED METHOD}\label{sec:methodology}
\begin{figure*}[!t]
		\centering
		\includegraphics[width=\linewidth]{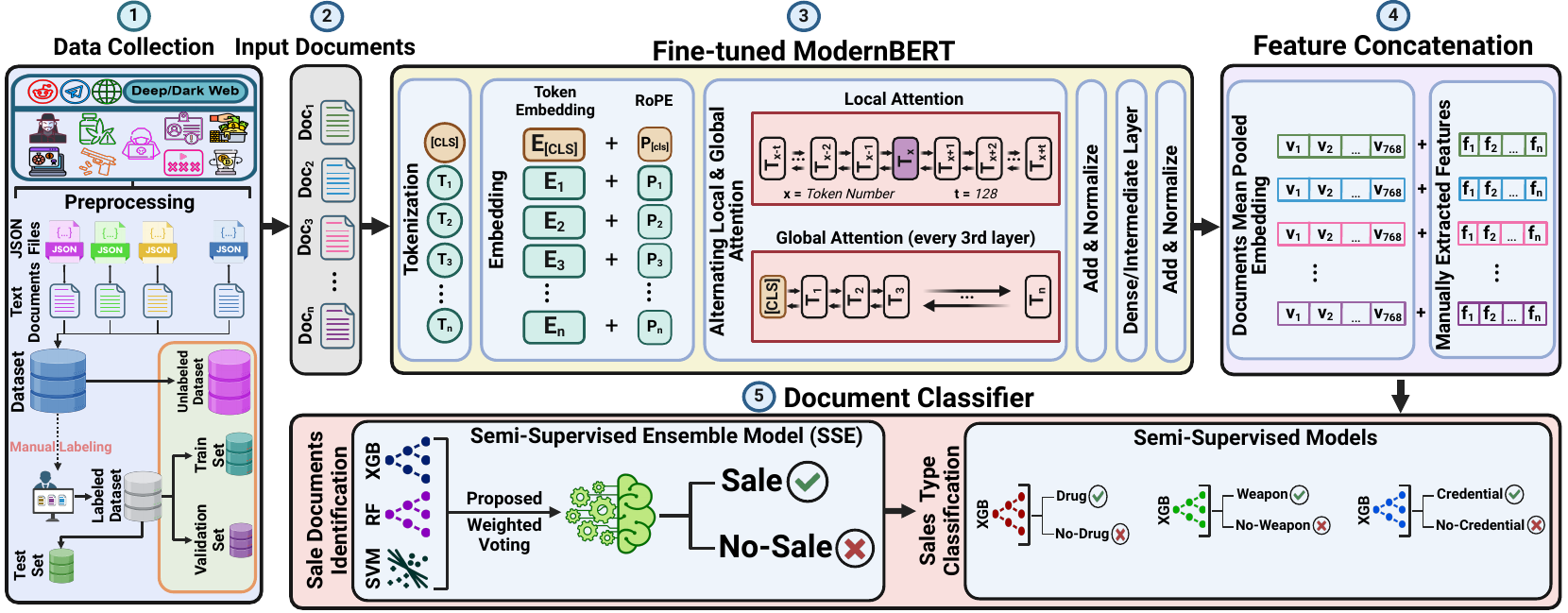}
		\caption{The proposed methodology overview}
		\label{fig:2}
	\end{figure*}

	This section outlines our proposed approach for detecting and classifying illicit online marketplace content. As illustrated in Fig.~\ref{fig:2}, the process begins with data collection using specialized crawlers that extract data from deep/dark web, Telegram, Reddit, and Pastebin. Each collected data sample is then converted into a document in a semi-structured JSON format, containing raw text and associated metadata, to standardize the data and facilitate subsequent analysis. Following this, a comprehensive preprocessing, document analysis, and feature extraction phase is conducted on each document to generate rich textual and structural representation. These representations serve as the foundation for training our sequential classification framework, which aims to accurately identify and categorize illicit marketplaces. The subsequent subsections provide a detailed explanation of each phase in the proposed pipeline.
	
	\subsection{Data Collection \& Preprocessing}
	As depicted in Fig.~\ref{fig:2}, the first stage involves an automated data collection system that gathers content from the deep/dark web, and platforms including Telegram, Reddit, and Pastebin. Starting from a list of seed URLs, the system expands through recursive crawling and spidering techniques, while utilizing proxies and a load balancer to efficiently navigate and retrieve HTTP-based content. To ensure uninterrupted data collection, the system is also equipped with mechanisms to bypass common access barriers, such as authentications and bot-detection systems, including CAPTCHAs. Following data collection, a duplicate filtering mechanism is applied to enhance the integrity and quality of the resulting dataset. Subsequently, the system extracts textual content from the collected data, while also retrieving textual metadata from the headers of non-textual content such as images, videos, and other multimedia files, capturing both the textual and contextual elements of the collected data. Next, the preprocessing begins, where the system uses Regular Expressions (RegEx) and HTML parsers to identify structured patterns like domain names, cryptocurrency addresses, and contact details. Finally, the parsed contents, along with the raw text extracted from each data sample, is converted into semi-structured JSON files, resulting in a coherent and standardized dataset ready for downstream analysis.
	
	\subsection{Feature Extraction}
	As shown in steps two to four of Fig.~\ref{fig:2}, feature extraction consists of two main components, comprising contextual embeddings generated from the fine-tuned ModernBERT, and a set of manually engineered features identified in our previous work \cite{yazdanjue2025cyber}. By combining these representations, the framework captures both the semantic and structural aspects of each document, thereby enhancing the model’s ability to accurately classify illicit marketplace content. The following subsections explain each component in detail.
	
	\subsubsection{Embeddings from Fine-Tuned ModernBERT}
	In the first component of feature extraction, we employ ModernBERT, a transformer-based LM optimized for processing lengthy documents, to generate deep, contextual embeddings \cite{12-warner2024smarter}. This model is particularly well-suited for our task, as many documents in our dataset, especially those from dark web forums and Reddit, often exceed typical 512-token limit of standard models such as BERT, ALBERT, and RoBERTa. To address this limitation, ModernBERT uses a sparse attention mechanism that allows it to handle sequences up to 8192 tokens, preserving long-range dependencies and capturing important contextual patterns of long texts. This sparse attention is realized through an alternating attention mechanism, which strategically combines global attention and local sliding window attention to balance computational efficiency and contextual depth. Every third layer in ModernBERT applies global attention, allowing each token to attend to every other in the sequence. This allows important context to be aggregated across the entire document, which is crucial for detecting illicit activity spread across multiple sections of lengthy documents. In contrast, the remaining layers use local sliding window attention, where each token only attends to a fixed number of surrounding tokens within a range of 128. This method reduces the quadratic complexity of full attention while preserving local coherence and contextual understanding. Moreover, ModernBERT integrates Rotary Positional Embeddings (RoPE), which encodes relative positional information to better preserve word order and long-range relationships without relying on fixed positional encodings. These mechanisms allow ModernBERT to process long documents without truncation, which is essential for our use case, where critical contextual patterns, such as identifiers, pricing details, or transaction instructions, may be scattered throughout the document.
	
	To further adapt ModernBERT to our application domain, we fine-tune it on collected data from the deep/dark web, Telegram, Reddit, and Pastebin. While ModernBERT is architecturally capable of processing long sequences, its pretrained parameters are based on general-purpose corpora and may not capture the unique linguistic characteristics, such as specialized jargon, slang, and patterns present in illicit marketplaces. Fine-tuning on the domain-specific dataset enables the model to adapt to these specialized characteristics, improving its ability to generate contextually rich embeddings aligned with the linguistic patterns in the mentioned sources. As illustrated in Fig.~\ref{fig:2}, the fine-tuned ModernBERT processes each document’s full text and produces a fixed-length embedding vector using the mean pooling technique. These embeddings effectively capture both the semantic meaning and the contextual depth of the documents, providing a robust input for the subsequent classification stages.
	
	\subsubsection{Manually Engineered Features}
	In the second feature extraction component, we manually extract features that complement the ModernBERT embeddings by capturing information not easily derived from text. Identified as the optimal feature set in our previous study \cite{yazdanjue2025cyber}, these features fall into three main categories: Layout Features, Pattern-Specific Item Features, and Metadata Features, each offering structural and contextual information that could be critical for accurately classifying illicit content. Each of these categories is detailed below.
	\begin{enumerate}[label=(\alph*)]
		\item Layout Features: These features capture structural and formatting aspects of documents that are not reflected in the text alone, offering 
        insights into the document’s content organization and visual structure, which can help identify illicit marketplace listings. Specifically, these extracted features include statistical properties of line widths (minimum, maximum, average, median, standard deviation, and variance), statistical properties of line indentation levels (using the same set of measures), and counts of non-empty and empty lines to evaluate the document’s structural density and formatting style.
			
		\item Pattern-Specific Item Features: These features capture important elements embedded within the document that cannot be fully understood by LMs. Using RegEx, we identify specific items such as Images, Credit Card Numbers, IP Addresses, Email Addresses, URLs, and Bitcoin Addresses, which could be crucial for detecting illicit transactions. For each item type (shown by “\textasteriskcentered”), two features are computed, including \textit{Item\_\textasteriskcentered\_Count}, representing the total number of occurrences of that item, and \textit{Item\_\textasteriskcentered\_Weight}, denoting the relative significance of the item by dividing its frequency by the total number of identified items, as formalized in Equation~\eqref{eq:1}.

        			\begin{equation}
				\label{eq:1}
				Item\_*\_Weight = \frac{Item\_*\_Count}{\text{Total Item Numbers}}
			\end{equation}

		\item Metadata Features: These features focus on the external contextual information about each document, such as its platform of origin and creation time, which can be critical for interpreting the document's role and relevance within illicit marketplaces. The extracted metadata features include Deep Web Source, Deep Web Source, Dark Web Source, Social Media Source, Document Date.

	\end{enumerate}
	
	\subsection{Proposed Sequential Classification Approach}
\begin{figure}[!t]
		\centering
		\includegraphics[width=.6\linewidth]{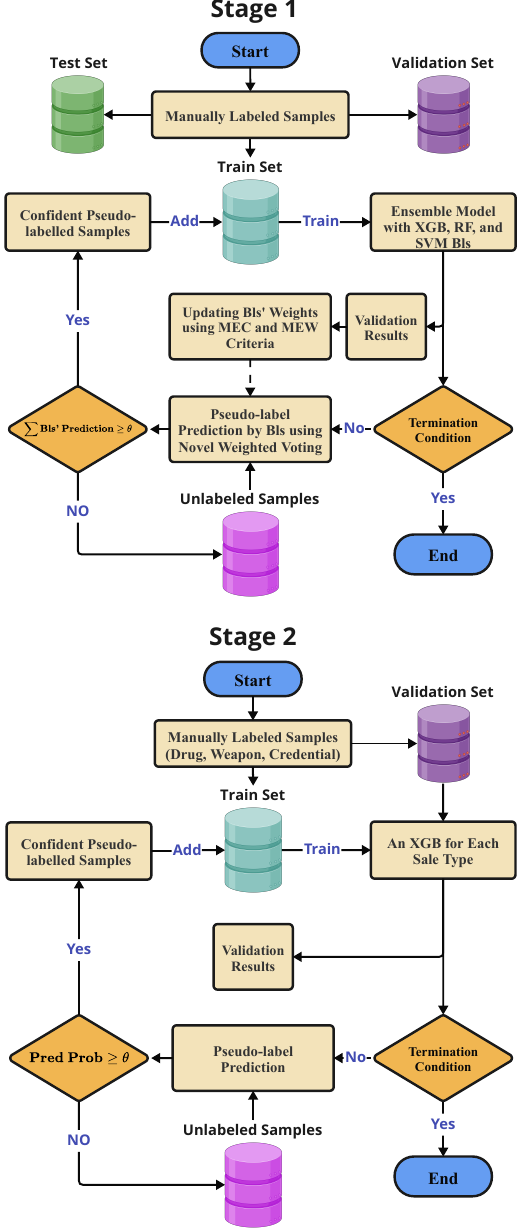}
		\caption{Flowchart of the proposed sequential classification framework. Stage 1 employs an SSE model comprising XGB, RF, and SVM with entropy-based voting to detect sales-related documents. Stage 2 uses three semi-supervised XGB classifiers to categorize them into drug, weapon, or credential labels.}
		\label{fig:3}
	\end{figure}

	In step 5 of Fig.~\ref{fig:2}, we introduce the proposed sequential document classification framework, designed to detect and categorize illicit marketplace content using a semi-supervised learning strategy. As illustrated in Fig.~\ref{fig:3}, the framework operates in two stages. The first stage identifies sales-related documents, while the second stage classifies them into three key categories, drug sales, weapon sales, and credential sales. The semi-supervised nature of this model allows it to effectively utilize a small expert-labeled dataset, divided into training, validation, and test sets, along with a large volume of unlabeled data to enhance the generalization and robustness of the classification model.
    
	%\begin{figure*}[!t]
	%	\centering
	%	\subfloat[\textbf{Stage 1: SSE Classifier}]{\includegraphics[width=2.5in]{A.pdf}%
		%	\label{fig:3a}}
	%	\hfil
	%	\subfloat[\textbf{Stage 2: Sales Type Classifier}]{\includegraphics[width=2.5in]{B.pdf}%
		%	\label{fig:3b}}
	%	\caption{Flowchart of the proposed sequential classification framework. Stage 1 employs an SSE model comprising XGB, RF, and SVM with entropy-based voting to detect sales-related documents. Stage 2 uses three semi-supervised XGB classifiers to categorize them into drug, weapon, or credential classes.
	%	\label{fig:3}
	%\end{figure*}
	
	As mentioned earlier, the first stage of the proposed sequential framework centers on the SSE model to identify sales-related documents across diverse sources, including the deep/dark web, Telegram, Reddit, and Pastebin. In this stage, the SSE model employs a self-training semi-supervised learning strategy that iteratively refines its predictions using both labeled and unlabeled data, while utilizing an ensemble of diverse BLs (XGB, RF, and SVM) combined through a novel weighted voting mechanism to enhance classification accuracy.
	
	After detecting sales-related documents with the SSE model, three distinct XGB-based semi-supervised models classify them into drug, weapon, and credential sales categories. The subsequent subsections provide a detailed explanation of each stage of the proposed classification model.
	
	\subsubsection{Sales Document Identification}
	As shown in Fig.~\ref{fig:3}, the SSE training process begins with a manually labeled dataset, where documents are annotated as either sales-related or non-sales. As previously noted, in the first stage of the sequential classifier, the SSE model is trained solely to distinguish between these two classes. The labeled data is then divided into training, validation, and test subsets, ensuring the SSE model is properly trained and validated during learning process and reliably evaluated during testing.

    Following this setup, an ensemble model is constructed comprising of three BLs (XGB, RF, and SVM). These BLs offer complementary strengths that improve the reliability and accuracy of the SSE model in detecting sales-related content in noisy, heterogeneous environments. XGB is highly efficient for handling large datasets and capturing complex patterns, making it particularly suitable for detecting complex sales-related documents. RF offers robustness through its ability to reduce overfitting by averaging multiple decision trees, ensuring stable predictions. Meanwhile, SVM excels in handling high-dimensional spaces and is effective for binary classification tasks, making it an excellent fit for distinguishing sales-related content.
	
	These BLs are initially trained on the labeled training subset. Once initial training is complete, the SSE model enters a self-training phase, during which pseudo-labels are assigned to unlabeled documents using the ensemble equipped with the proposed weighted voting strategy. Through multiple iterations, confidently pseudo-labeled samples are incorporated into the training set, progressively expanding the training data pool and enhancing the model’s ability to distinguish sales-related content more effectively. To control the self-training loop and prevent overfitting, the SSE model applies three termination criteria to decide when to stop the iterative labeling process. The process is stopped if the maximum number of iterations has been reached, if all unlabeled samples have been pseudo-labeled, or if no confident predictions can be made for the remaining samples. If none of these conditions are met, the model proceeds to predict pseudo-labels for the remaining unlabeled samples.
	
	A distinguishing feature of the proposed SSE model is its novel weighted voting strategy, which enhances decision reliability by assigning greater influence to BLs with higher prediction confidence. Instead of treating all outputs equally, this mechanism dynamically adjusts each BL’s weight based on its predictions’ certainty. By prioritizing confident predictions and down-weighting uncertain ones, this voting strategy refines the ensemble decisions, ultimately improving the accuracy and robustness of sales-related content detection.
    
    At the start of training, all BLs in the ensemble (XGB, RF, and SVM) are assigned equal weights (i.e., 0.33). As training progresses, however, the model dynamically updates these weights using two entropy-based criteria, $MEC$ and $MEW$. These criteria are computed for each BL using its performance on the validation set and updated at each iteration. By promoting BLs that demonstrate high confidence in correct predictions (low $MEC$) and low confidence in incorrect ones (high $MEW$), this mechanism ensures that more reliable classifiers exert greater influence on the final ensemble decision. The $MEC$ criterion measures the average uncertainty associated with a BL’s correctly classified samples. A lower $MEC$ value indicates that the classifier is not only accurate but also confident in those correct predictions, producing less entropy in these cases. The $MEC$ for a given BL is calculated using Equation ~\eqref{eq:2}:
	\begin{equation}\label{eq:2}
		MEC = \frac{1}{N_C} \times \sum_{i=1}^{N_C} H(p_{c_i})
	\end{equation}
	where $N_C$ is the number of correct predictions, and $H(p_{c_i})$ is the entropy of each correct prediction.
	
	In parallel, the $MEW$ measures the average uncertainty associated with a BL’s incorrect classifications. A higher $MEW$ value is preferred, as it indicates that the classifier exhibits greater hesitation or uncertainty when making incorrect predictions, an important characteristic for minimizing the impact of unreliable outputs in the ensemble. The $MEW$ for a given BL is computed using Equation~\eqref{eq:3}:
	\begin{equation}\label{eq:3}
		MEW = \frac{1}{N_W} \times \sum_{i=1}^{N_W} H(p_{w_i})
	\end{equation}
	where $N_W$ is the number of incorrect predictions, and $H(p_{w_i})$ is the entropy of each incorrect prediction.
	
	It is worth noting that the entropy $H(p)$ for each prediction is calculated using Shannon’s entropy formula \cite{32-shannon1948mathematical}, shown in Equation~\eqref{eq:4}.
	\begin{equation}\label{eq:4}
		H(p) = -\sum_{i=1}^{n} p_i \times \log(p_i)
	\end{equation}
	where $n$ is the number of classes, and $p_i$ is the predicted probability of sample $i$.
	
	Building on the calculated $MEC$ and $MEW$ values, the SSE model updates the weight of each BL using a dynamic weighting strategy. Specifically, the weight assigned to each classifier is computed as the ratio of its $MEW$ to $MEC$, reflecting its reliability in terms of both confident correct predictions and cautious incorrect ones. To ensure that all weights are proportionally balanced, the ratio is normalized across all BLs in the ensemble. This process is formalized in Equation~\eqref{eq:5}:
	\begin{equation}\label{eq:5}
		w_i = \frac{\left( \frac{MEW_i}{MEC_i} \right)}{\sum_{i=1}^{N_{BL}} \frac{MEW_i}{MEC_i}}
	\end{equation}
	where $N_{BL}$ represents the number of BLs. This weighting strategy ensures that classifiers demonstrating high confidence in their correct predictions (reflected by lower $MEC$) and increased uncertainty when making incorrect predictions (reflected by higher $MEW$) are given proportionally more influence in the ensemble's final decision. In essence, the mechanism prioritizes classifiers that are both confidently correct and cautiously wrong, thereby improving the reliability of pseudo-label assignments and enhancing overall model accuracy during self-training.
	
	Once the weights for each BL are dynamically updated using the $MEC$ and $MEW$ criteria, the SSE model proceeds to assign pseudo-labels to the unlabeled samples. For each unlabeled sample, the class probability predictions from the BLs are aggregated using the updated weights to compute the total prediction probability ($TPP$) for each class. This process enables the ensemble to form a weighted consensus that accurately reflects the relative reliability and predictive confidence of each individual classifier. The $TPP$ for class $c$ is calculated using Equation~\eqref{eq:6}:
	\begin{equation}\label{eq:6}
		TPP_c = \sum_{i=1}^{N_{BL}} w_i \times cpp_{i}
	\end{equation}
	Where $TPP_c$ is the total prediction probability for class $c$, either “sale” or “no-sale”, $N_{BL}$ is the number of base learners in the ensemble, $w_i$ represents the weight assigned to the $i$th classifier calculated by Equation~\eqref{eq:5}, and $cpp_i$ is the class prediction probability for the $i$th classifier, indicating the likelihood that a given sample belongs to class $c$. . At its core, the $TPP$ is a reliability-weighted consensus of the BLs, with each classifier’s contribution scaled by its confidence via the $MEW$-to-$MEC$ ratio.
	
	 After computing the $TPP$ values for both classes, “sale” and “no-sale”, the SSE model assigns a pseudo-label to the unlabeled sample based on the class with the higher total prediction probability. This decision rule is formalized in Equation~\eqref{eq:7}:
	\begin{equation}\label{eq:7}
		PI = 
		\begin{cases}
			\text{sale} & \text{if } TPP_{sale} > TPP_{no\_sale} \\
			\text{no\_sale} & \text{else } TPP_{no\_sale} > TPP_{sale}
		\end{cases}
	\end{equation}
	
	Once the pseudo-label is determined based on the higher total prediction probability to the unlabeled sample, the model calculates a confidence score by averaging the predicted probabilities of the assigned class across all BLs, expressed in Equation~\eqref{eq:8}. If this score exceeds a predefined confidence threshold $\theta$, the sample is considered confidently labeled. For instance, if an unlabeled sample is pseudo-labeled as “sale” and the average predicted probability for the “sale” class across all BLs meets or exceeds the $\theta$ value, the sample is considered confidently labeled.
	\begin{equation}\label{eq:8}
		CS_{Pl} = \frac{1}{N_{BL}} \sum_{i=1}^{N_{BL}} cpp_{Pl,i}
	\end{equation}
	In Equation~\eqref{eq:8}, $CS_{Pl}$ is the confidence score for the pseudo-label “Pl” (either “sale” or “no-sale”). Also, $cpp_{Pl,i}$ is the class prediction probability from the $i$th base learner for the predicted pseudo-label ($Pl$).
	
	Confidently labeled samples are then incorporated into the training set to enrich the learning process. By adding only high-confidence pseudo-labeled samples, the model minimizes the risk of introducing noise and ensures that each retraining iteration is grounded in reliable data. The SSE model is subsequently retrained on this updated training set, which now includes both the original labeled samples and the newly confident pseudo-labeled samples. This process of prediction, confident pseudo-labeling, and retraining continues iteratively until one of the predefined termination criteria, such as reaching the maximum number of iterations, no unlabeled samples remaining in the unlabeled pool, or failing to identify confident samples, is met. Through this cycle, the model progressively refines its ability to distinguish sales-related documents with greater robustness and accuracy.
	
	\subsubsection{Classifying Sale Document Types}
	In the second stage of our proposed framework, illustrated in Fig.~\ref{fig:3}, we implement three distinct self-training semi-supervised classifiers, each based on XGB, to further categorize the documents identified as sales-related documents at stage one, into specific illicit trade types (drug, weapon, and credential sales). XGB is selected as the classifier for the second stage due to its strong performance in the first stage, demonstrating its effectiveness in document classification tasks. Also, using a single XGB model per category simplifies the system architecture, reduces computational overhead and accelerates training. Moreover, given the smaller sample sizes for each category (drug, weapon, and credential), using individual XGB models helps mitigate the risk of overfitting and promotes better generalization without capturing noise or anomalies in the data.
    
    The training data for this stage is derived from the same manually labeled dataset used in the first stage. While the first stage used only the sale label to detect sales-related content, this stage shifts focus to the remaining three labels, i.e., drug, weapon, and credential sales. For each XGB classifier, the subset of samples labeled with its corresponding category (e.g., drug sales for the drug classifier) is extracted from the original training data and split into separate training set (80\%) and validation set (20\%), enabling category-specific training and evaluation.
	
	Similar to the first stage, this second stage adopts a self-training semi-supervised approach to utilize both labeled and unlabeled data. The unlabeled samples used here are those documents that were pseudo-labeled as “sale” by the SSE model in stage one, though their specific category remains unknown. Each unlabeled sample is independently evaluated by all three XGB classifiers, each generating a prediction probability for its respective category: drug sales, weapon sales, or credential sales. For instance, if the drug XGB classifier assigns a probability above the confidence threshold ($\theta$), the sample is pseudo-labeled as ‘drug’ and added to its training set; otherwise, it remains unlabeled for future refinement in subsequent self-training iterations.
	After adding the confidently pseudo-labeled samples to the training set of the corresponding XGB classifier, that is subsequently retrained using this expanded dataset. This iterative self-training process continues, with classifiers gradually labeling and learning from new high-confidence samples in each cycle, until a predefined stopping criterion is met. Once the training process for all three XGB classifiers in the second stage is complete, the proposed sequential classification framework enters the testing phase to evaluate its accuracy and generalization capabilities on unseen data.
	
	\section{EXPERIMENTAL EVALUATION}\label{sec:exset}
	To evaluate the effectiveness of the proposed sequential document classification framework, we conducted a comprehensive set of experiments under various conditions and benchmarked its performance against several existing approaches. All experiments were implemented in Python 3.12 and executed on a Google Colab environment equipped with an NVIDIA A100 GPU (40 GB VRAM), using essential packages, such as Transformers, scikit-learn, and XGBoost. In the following subsections, we first introduce our collected dataset, along with the hyperparameter settings and evaluation metrics employed to evaluate the results. We then begin our experimental analysis by comparing the impact of traditional (TF-IDF, Doc2Vec and GloVe) and transformer-based (pretrained and fine-tuned $\text{BERT}_{\text{base}}$, $\text{ALBERT}_{\text{base}}$, Longformer-4096, BigBird-4096, and ModernBERT) text representations on the performance of the proposed classification framework. To further support this comparison, we performed a ranking-based statistical analysis using the Friedman test to evaluate the performance differences among 12 models across three evaluation metrics. We then assess the robustness of our suggested sequential classification framework under different proportions of labeled data, highlighting its semi-supervised learning capabilities. Next, we evaluate the effectiveness of our model by comparing each of its two stages against a range of supervised and semi-supervised baselines. In the first stage, we assess the SSE model’s ability to detect sales-related content by comparing it with supervised models (XGB, RF, SVM, and their ensemble variants with majority voting and the proposed weighted voting strategies) and their semi-supervised counterparts. In the second stage, the three semi-supervised XGB classifiers are compared to both supervised and semi-supervised versions of XGB, RF, and SVM for category-specific classification (drug, weapon, and credential categories). Finally, we benchmark our model’s generalizability against BERT \cite{9-devlin2019bert}, ALBERT \cite{10-lan2019albert}, Longformer \cite{33-beltagy2020longformer}, BigBird \cite{34-zaheer2020big}, ModernBERT \cite{12-warner2024smarter} and DarkBERT \cite{29-jin2023darkbert} across multiple datasets and settings.
	
	\subsection{Dataset Overview}
	As explained before, to support our classification experiments, we constructed a dataset by collecting 21,575 samples from various sources, including the deep/dark web, Telegram, Reddit, and Pastebin, covering the period from September 2021 to September 2023. Out of the 21,575 collected samples, 1,575 (approximately 7\%) were manually annotated by domain experts to form the labeled subset used for training, validation, and testing. To ensure a consistent evaluation framework, the labeled data was split into 60\% for training (945 samples), 20\% for validation (315 samples), and 20\% for testing (315 samples). The test set was held out and remained unseen throughout the training process, serving as a fixed benchmark to evaluate the generalization performance of all models. The remaining 20,000 samples were used as unlabeled data to support semi-supervised learning process. Table~\ref{tab:2} presents the distribution of both labeled and unlabeled samples across different data sources.
    
	\begin{table}[!t]
		\centering
		\caption{The number of samples from various sources in the dataset}
		\label{tab:2}
		\resizebox{\columnwidth}{!}{%
			\begin{tabular}{|c|cccc|}
				\hline
				\multirow{2}{*}{\textbf{Datasets}} & \multicolumn{4}{c|}{\textbf{Data Sources}} \\ \cline{2-5}
				& \multicolumn{1}{c|}{\textbf{Deep Web}} & 
				\multicolumn{1}{c|}{\textbf{Dark Web}} & 
				\multicolumn{1}{c|}{\textbf{\begin{tabular}[c]{@{}c@{}}Social Media\\ (Telegram \& Reddit)\end{tabular}}} & 
				\textbf{Pastebin} \\ \hline
				Labeled & \multicolumn{1}{c|}{450} & \multicolumn{1}{c|}{825} & \multicolumn{1}{c|}{200} & 100 \\ \hline
				Unlabeled & \multicolumn{1}{c|}{4382} & \multicolumn{1}{c|}{13107} & \multicolumn{1}{c|}{1899} & 612 \\ \hline
			\end{tabular}%
		}
	\end{table}
	To further illustrate the composition of the labeled dataset, Table~\ref{tab:3} presents the number of samples assigned to each specific category (i.e., sale, drug sale, weapon sale, and credential sale. For each category, samples are grouped into ‘Yes’ (samples that belong to the label) and ‘No’ (samples that do not), reflecting the binary labeling used for classification.
	\begin{table}[!t]
		\centering
		\caption{The number of samples related to each class of each label in the manually labeled dataset}
		\label{tab:3}
		\resizebox{\columnwidth}{!}{%
			\begin{tabular}{|c|cc|cc|cc|cc|}
				\hline
				\textbf{Dataset} &
				\multicolumn{8}{c|}{\textbf{Labels}} \\ \cline{2-9}
				&
				\multicolumn{2}{c|}{\textbf{Sale}} &
				\multicolumn{2}{c|}{\textbf{Drug Sale}} &
				\multicolumn{2}{c|}{\textbf{Weapon Sale}} &
				\multicolumn{2}{c|}{\textbf{Credential Sale}} \\ \cline{2-9}
				& \textbf{Yes} & \textbf{No} & \textbf{Yes} & \textbf{No} & \textbf{Yes} & \textbf{No} & \textbf{Yes} & \textbf{No} \\ \hline
				\textbf{Manually Labeled Set} & 826 & 749 & 365 & 461 & 398 & 428 & 356 & 470 \\ \hline
			\end{tabular}%
		}
	\end{table}
	
	\subsection{Model Hyperparameters}
	To optimize the performance of our proposed classification model, we conduct a comprehensive grid search to tune hyperparameter for fine-tuning the ModernBERT model, training the BLs of the proposed SSE model, as well as configuring the three semi-supervised XGB models in the second stage. A summary of search ranges and the optimal values is provided in \ref{tab:4}. It is noteworthy that for ModernBERT, we explored various configurations, evaluating learning rates [1e-5,3e-5,5e-5,8e-5] and epoch sizes [5, 10, 15, 20]. 
    The model was fine-tuned using the AdamW optimizer with a batch size of 10, a gradient accumulation of 3 to manage memory requirements, and a weight decay of 5e-6. The grid search results identified the optimal learning rate as 5e-5 and the optimal number of epochs as 15. Also, for the BLs in the SSE model (XGB, RF, and SVM) and the three XGB classifiers in the second stage, we explored key hyparameters, such as learning rate, number of estimators, maximum depth, subsample ratio, regularization terms, and kernel settings. Additionally, confidence thresholds and maximum iteration limits were tuned in both stages to ensure robust and reliable pseudo-labeling.
	\begin{table}[!t]
		\centering
		\caption{The hyperparameter ranges and selected values}
		\label{tab:4}
		\resizebox{\columnwidth}{!}{%
			\begin{tabular}{|c|c|c|c|}
				\hline
				\textbf{Models} & \textbf{Hyperparameters} & \textbf{Search Range} & \textbf{Optimal Value} \\
				\hline
				
				\multirow{2}{*}{ModernBERT} 
				& Learning Rate & [1e-5, 3e-5, 5e-5, 8e-5] & 5e-5 \\
				& Epoch Size    & [5, 10, 15, 20]           & 10 \\
				\hline
				
				\multirow{6}{*}{$\text{XGB}_{\text{(SSE, Drug, Weapon, Credential)}}$} 
				& Learning Rate     & [0.01, 0.05, 0.1, 0.2]               & 0.1, 0.1, 0.05, 0.1 \\
				& N-estimators      & [100, 200, 300, 400, 500]           & 400, 300, 300, 500 \\
				& Max Depth         & [3, 5, 7, 10]                        & 5, 5, 5, 7 \\
				& Subsample         & [0.6, 0.7, 0.8]                      & 0.7, 0.6, 0.6, 0.7 \\
				& L1 Regularization & [0.01, 0.05, 0.1]                    & 0.01, 0.01, 0.01, 0.01 \\
				& L2 Regularization & [0.01, 0.05, 0.1]                    & 0.01, 0.01, 0.01, 0.01 \\
				\hline
				
				\multirow{4}{*}{$\text{RF}_{\text{SSE}}$}
				& Max Depth     & [3, 5, 7, 10]                  & 5 \\
				& N-estimators  & [100, 200, 300, 400, 500]      & 400 \\
				& Max Features  & [Auto, Sqrt]                   & Auto \\
				& Split Criterion & [Gini, Entropy]             & Gini \\
				\hline
				
				\multirow{3}{*}{$\text{SVM}_{\text{SSE}}$}
				& Kernel          & [Linear, Sigmoid, RBF]            & RBF \\
				& Regularization (\(C\)) & [0.001, 0.01, 0.1, 1, 10] & 0.01 \\
				& Gamma           & [0.001, 0.01, 0.1, 1, 10]         & 0.1 \\
				\hline
				
				\multirow{2}{*}{$\text{Self-training}_{\text{(SSE, Drug, Weapon, Credential)}}$}
				& \(\theta\)           & [0.8, 0.85, 0.9, 0.95]              & 0.9, 0.9, 0.85, 0.9 \\
				& Max Iteration  & [25, 50, 75, 100]                   & 75, 25, 25, 50 \\
				\hline
				
			\end{tabular}%
		}
	\end{table}
	
	\subsection{Performance Metrics}
	To evaluate the performance of the compared classification models, we employ three core metrics, including Accuracy (Acc), F1-score (F1), and Matthews Correlation Coefficient (MCC), defined in the Equations~\eqref{eq:Acc} to~\eqref{eq:MCC}, respectively.
      
      \begin{equation}\label{eq:Acc}
      \text {Accuracy} = \frac{(TP + TN)}{(TP + TN + FP + FN)} 
      \end{equation}

       \begin{equation}\label{eq:F1}
        	\text{F1 Score} = 2 \times \frac{\text{Precision} \times \text{Recall}}{\text{Precision} + \text{Recall}}
        \end{equation}

    {\small
			\begin{equation}\label{eq:MCC}
				\text{MCC} = \frac{(TP \times TN) - (FP \times FN)}{\sqrt{(TP + FP)(TP + FN)(TN + FP)(TN + FN)}}
			\end{equation}
}
		In these equations, $TP$, $TN$, $FP$, and $FN$ represent the counts of true positives, true negatives, false positives, and false negatives, respectively. Also, F1-score is the harmonic mean of precision defined as $\text TP / (TP + FP)$ and recall defined as $\text TP / (TP + FN)$, capturing the balance between correct positive predictions and the proportion of actual positive cases detected. This metric is especially valuable for class-imbalanced datasets. Moreover, MCC measures the quality of binary classifications, providing a balanced evaluation even with imbalanced classes. MCC is calculated using Equation~\eqref{eq:MCC}. MCC values range from -1 to +1, and to scale this range between 0 and 1, we apply a simple linear transformation, as illustrated in Equation~\eqref{eq:TMCC}. This transformation maps the MCC range from [-1, 1] to [0, 1], yielding a metric referred to as the Transformed Matthews Correlation Coefficient (TMCC). In this transformed scale, a TMCC value of 1 denotes perfect prediction, 0.5 suggests no better than the random prediction, and 0 indicates complete disagreement between the predicted and ground truth.
        
		\begin{equation}\label{eq:TMCC}
			\text{TMCC} = \frac{\text{MCC} + 1}{2}
		\end{equation}
	
	\subsection{Comparison of Text Representation Techniques for the Proposed Classification Framework}
	To identify the most effective text representation strategy for our classification framework, we conducted a comprehensive evaluation of both traditional and transformer-based techniques. Traditional approaches included TF-IDF, Doc2Vec, and GloVe, while the transformer-based LMs involved contextual embeddings from $\text{BERT}_{\text{base}}$, $\text{ALBERT}_{\text{base}}$, Longformer-4096, and BigBird-4096, and ModernBERT. Each transformer model was evaluated in both its pretrained and fine-tuned forms. To ensure fair comparison, all fine-tuned models were trained on the same dataset. Besides, in alignment with their original studies [9, 10, 33, 34], we used model-specific learning rates, setting them to 2e-5 for $\text{BERT}_{\text{base}}$, 1e-5 for $\text{ALBERT}_{\text{base}}$, and 3e-5 for Longformer-4096 and BigBird-4096.
	
\begin{figure}[!t]
		\centering
		\includegraphics[width=\linewidth]{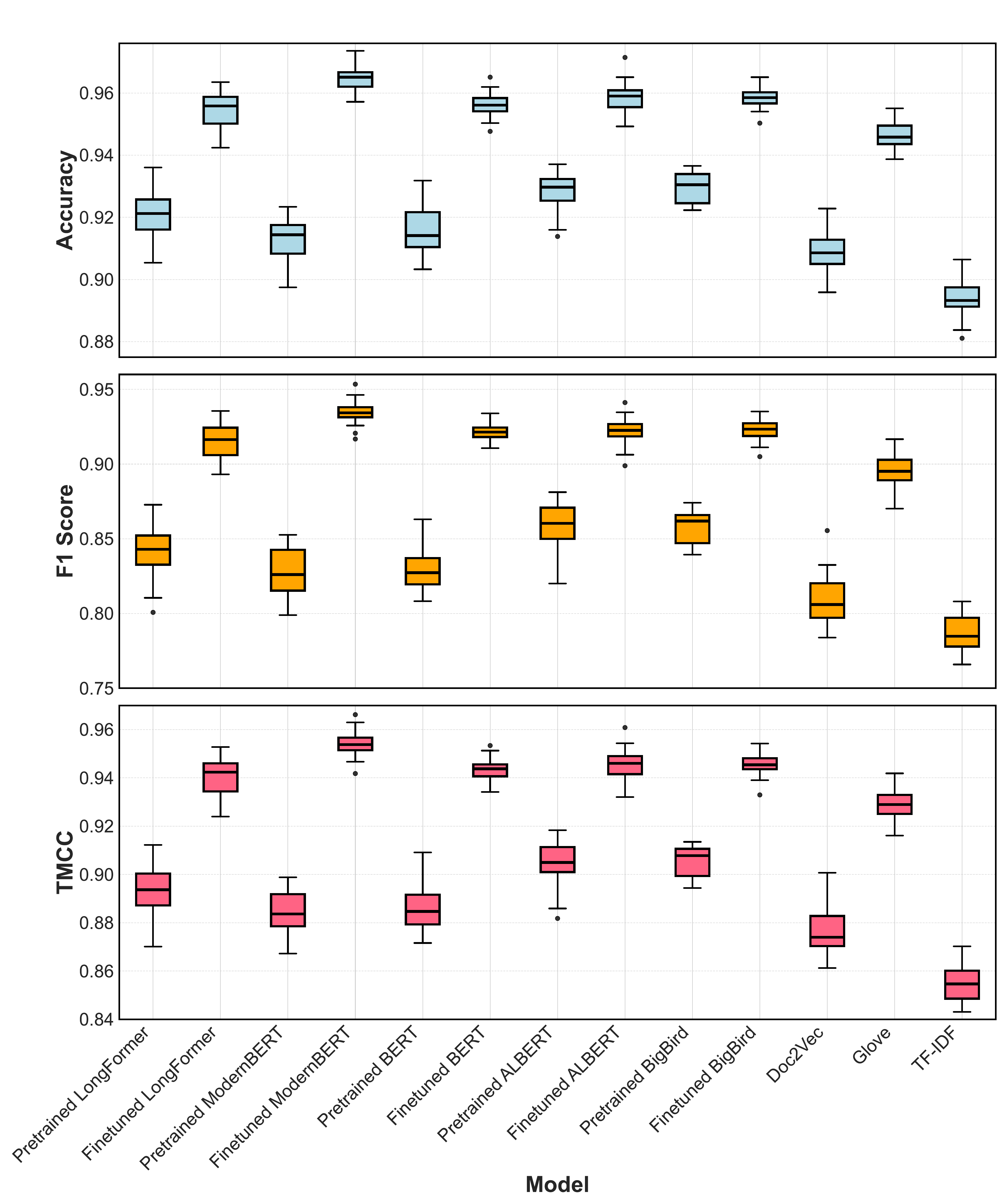}
		\caption{Performance comparison of text representation techniques based on macro accuracy, macro F1 Score, and macro TMCC score over 30 runs.}
		\label{fig:4}
	\end{figure}

	Also, to create a unified input representation for our model, we concatenated the embeddings generated by each method with our manually engineered features. The full classification pipeline was then executed 30 times per input representation to ensure robust evaluation. Given that our task involves a multi-label classification task with four categories (sale, drug, weapon, and credential), we utilized macro-averaged Accuracy, F-score, and TMCC metrics to ensure a balanced assessment. The results, visualized as boxplots in Fig.~\ref{fig:4}, illustrate variability and stability for each representation method over 30 runs.
	
	As illustrated in Fig.~\ref{fig:4}, fine-tuned transformer-based models clearly outperform other representation methods across all three evaluation metrics (Accuracy, F1-score, and TMCC), demonstrating their superiority in capturing domain-specific nuances. Among them, fine-tuned ModernBERT outperforms all others, delivering the highest average performance with minimal variance across 30 runs, highlighting its robustness and stability. In comparison, the performance of the classification model using the pretrained variants of transformer models show noticeably lower performance, emphasizing the critical role of domain-specific fine-tuning in adapting to the unique linguistic characteristics of content from the deep/dark web, Telegram, Reddit, and Pastebin. Using traditional approaches, including TF-IDF, Doc2Vec, and GloVe, generally exhibit lower overall performance compared to fine-tuned transformer models. However, GloVe demonstrates relatively competitive performance, highlighting its relative strength in capturing some aspects of textual structure. Despite this, traditional methods tend to show lower accuracy with wider interquartile ranges, reflecting their limited robustness in modeling the complex and obfuscated language of illicit marketplace content.
	
	A similar trend is observed in the F1-scores, where fine-tuned transformer-based embeddings used as part of the input to our classification framework consistently outperform other representation methods. Fine-tuned ModernBERT once again stands out, achieving some of the highest and most stable F1-scores, which highlights its effectiveness in modeling the complex and nuanced language typical of illicit marketplace content across the evaluated platforms. This superiority is further confirmed by the TMCC results, where the classification framework using ModernBERT embeddings consistently achieves high scores, demonstrating reliable performance in distinguishing between classes with balanced true and false prediction rates.
	
	To complement the quantitative performance comparison, we conducted a ranking-based statistical analysis using the Friedman test to further validate these findings. For each of the 30 independent runs, the models were ranked separately based on their macro Accuracy, F1-score, and TMCC, with the best-performing algorithm assigned a rank of 1 and the worst-performing algorithm assigned a rank of 13. To compute a global ranking that accounts for all three metrics simultaneously, the overall rank of each algorithm was calculated by averaging its mean ranks across Accuracy, F1-score, and TMCC. 

	\begin{table}[!t]
		\centering
		\caption{Overall Ranking and Mean Ranks of Representation Techniques Across Accuracy, F1 Score, and TMCC.}
		\label{tab:5}
		\resizebox{\columnwidth}{!}{%
			\begin{tabular}{|c|c|c|c|c|}
				\hline
				\textbf{Representation Techniques} & \textbf{Accuracy} & \textbf{F1 Score} & \textbf{TMCC} & \textbf{Overall Ranking} \\
				\hline
				TF-IDF & 12.90 & 12.93 & 12.97 & 12.93 \\
				Doc2Vec & 11.53 & 11.77 & 11.60 & 11.63 \\
				Pretrained ModernBERT & 10.57 & 10.27 & 10.47 & 10.43 \\
				Pretrained BERT & 10.13 & 10.37 & 10.07 & 10.19 \\
				Pretrained Longformer & 9.40 & 9.30 & 9.50 & 9.40 \\
				Pretrained ALBERT & 7.83 & 7.60 & 7.73 & 7.72 \\
				Pretrained BigBird & 7.53 & 7.77 & 7.67 & 7.66 \\
				GloVe & 5.77 & 5.97 & 5.87 & 5.87 \\
				Fine-tuned Longformer & 4.40 & 4.30 & 4.37 & 4.36 \\
				Fine-tuned BERT & 3.80 & 3.47 & 3.67 & 3.64 \\
				Fine-tuned ALBERT & 2.83 & 3.10 & 3.03 & 2.99 \\
				Fine-tuned BigBird & 3.03 & 3.03 & 2.90 & 2.99 \\
				Fine-tuned ModernBERT & 1.07 & 1.13 & 1.17 & 1.12 \\
				\hline
			\end{tabular}
		}
	\end{table}
	
	As shown in Table~\ref{tab:5}, the ranking analysis confirms that combining embeddings from fine-tuned ModernBERT with manually engineered features enables our classification model to achieve the best overall performance, with a top-ranked average score of 1.12 across all evaluation metrics. In contrast, traditional techniques such as TF-IDF and Doc2Vec receive the worst overall rankings, reflecting their limited capability in capturing the complex and context-dependent language present in illicit marketplace content. To assess the statistical significance of the observed differences, we applied the Friedman test, which produced a test statistic of 337.59 and a p-value of 5.81E-65. Since the p-value is far below the significance threshold of 0.05, we reject the null hypothesis, confirming that at least one algorithm performs significantly differently from the others. These findings provide strong statistical evidence that fine-tuned transformer models, when integrated into the feature extraction pipeline of our classification framework, achieve significantly higher rankings than both traditional representation techniques and pretrained models. Due to its consistently superior and stable performance across all evaluation metrics, fine-tuned ModernBERT was selected as the primary embedding method, used alongside manually engineered features in the subsequent experiments.
	
	\subsection{The Proposed Model Performance with Varying Labeled Data Sizes}
	To further assess the robustness and adaptability of the proposed model, we evaluate its performance across varying proportions of labeled data. This experiment highlights the model’s capability to effectively utilize both limited labeled data and a large pool of unlabeled samples within the sequential semi-supervised learning framework. Specifically, in this experiment, we consider  5\%, 15\%, 25\%, 50\%, 75\%, 90\%, and 100\% of the labeled data as training data subsets, each sampled randomly from the full labeled set. Each labeled subset is then combined with the full set of 20,000 unlabeled samples during training. Also, throughout training, the previous fixed validation set is used to compute entropy-based weights for each BL in the ensemble, supporting the SSE model’s weighted voting mechanism. Following training with each labeled subset combined with the full unlabeled set, the model is evaluated on the fixed, unseen test set using macro-averaged Accuracy, F1 Score, and TMCC.
	
\begin{figure}[!t]
		\centering
		\includegraphics[width=\linewidth]{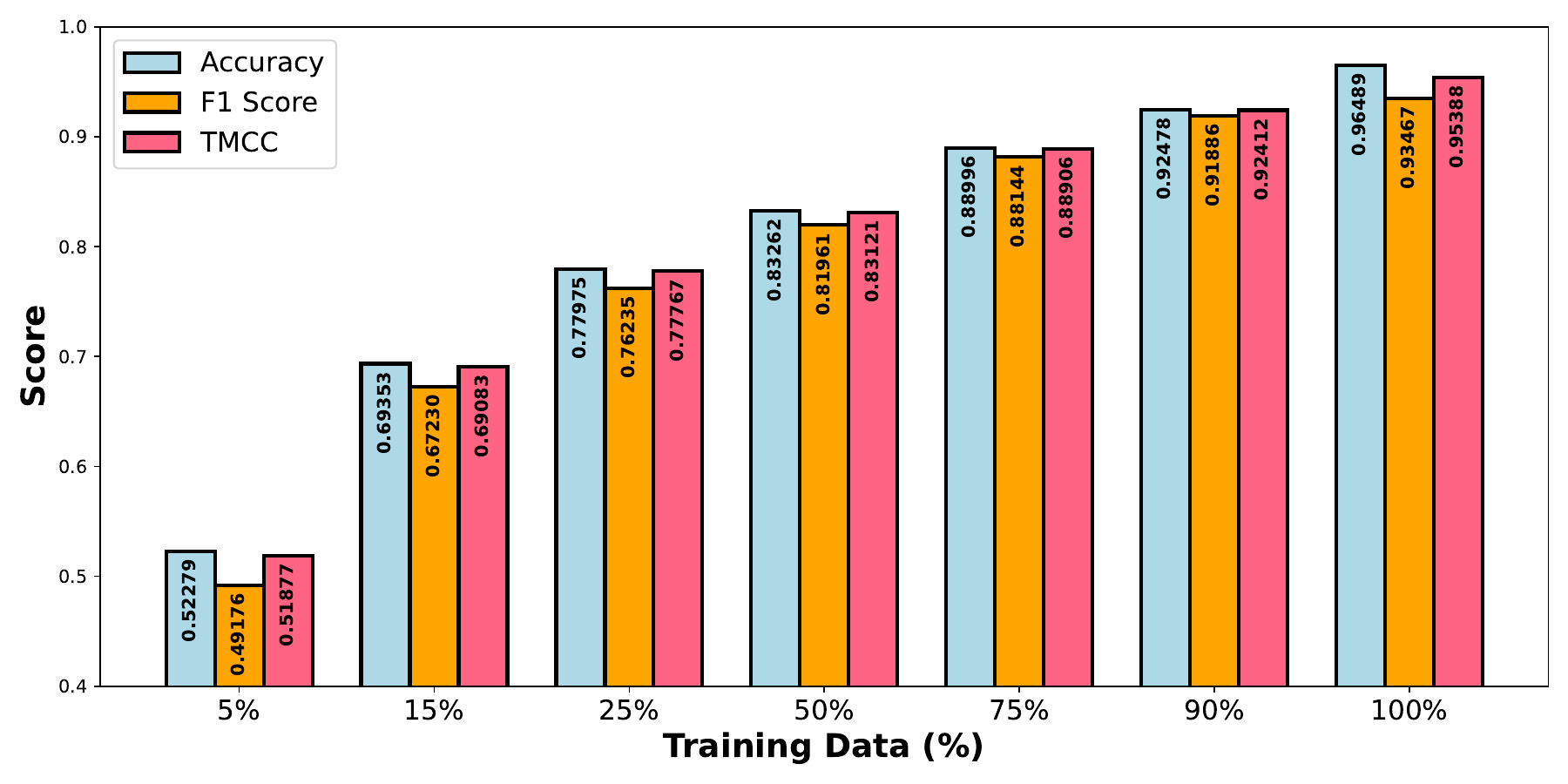}
		\caption{Performance of SSE Model with Varying Labeled Data Sizes Across Key Metrics}
		\label{fig:5}
	\end{figure}

	As shown in Fig.~\ref{fig:5}, there is a consistent and clear improvement across all three metrics as the proportion of labeled data increases. At just 5\% of the labeled training data, the model achieves Accuracy of 0.52279, F1 Score of 0.49176, and TMCC of 0.51877. These results indicate the model’s ability to learn useful patterns even from minimal labeled training set, thanks to its semi-supervised self-training mechanism and ensemble learning strategy. As the proportion of labeled training data increases to 15\% and 25\%, the model’s performance improves notably across all evaluation metrics. At 25\% of the labeled training data, the model achieves an Accuracy of 0.77975 and an F1 Score of 0.76235, representing more than a 27\% improvement in F1 Score compared to to the 5\% setting. This gain underscores the model’s ability to quickly benefit from additional labeled data, while continuing to exploit the large pool of unlabeled samples for enhanced generalization. At the 50\% of the labeled training data, the model surpasses the 0.83 range across all evaluation metrics, demonstrating a high level of reliability with just half of the labeled training data. Specifically, the TMCC reaches 0.83121, reflecting the model’s ability to make stable predictions across multiple categories. As the proportion of labeled training data increases beyond 75\%, the performance curve begin to plateau, indicating diminishing returns. Between 90\% and 100\% of the labeled training data, the F1 Score improves marginally by only 1.6 percentage points (from 0.91886 to 0.93467), and TMCC increases slightly from 0.92412 to 0.95388. 
    These findings confirm the robustness and practicality of the proposed model, even with a small portion of the labeled training data. With just 25–50\% of the labeled training data, the model demonstrates strong and reliable performance, due to its sequential classification design and the integration of semi-supervised learning with a novel entropy-based weighted voting ensemble learning. This capacity to effectively utilize large volumes of unlabeled data makes the model particularly well-suited for real-world applications, where labeled data is scarce, costly, or time-consuming to obtain.
    
	\subsection{Comparative Analysis of Proposed Model With Other Classification Methods}
	To gain deeper insights into the effectiveness of the proposed sequential document classification framework, we conduct a detailed comparison of each of its two stages against various semi-supervised and supervised models. We begin by evaluating the first stage (SSE model), focusing on its ability to detect sales-related documents and indicating the benefits of using semi-supervised learning. The models in this comparison are grouped into two categories: semi-supervised and supervised. The semi-supervised group consists of semi-supervised XGB ($\text{SS}_{\text{XGB}}$), semi-supervised RF ($\text{SS}_{\text{RF}}$), semi-supervised SVM ($\text{SS}_{\text{SVM}}$), and a semi-supervised ensemble with a majority voting strategy ($\text{SSE}_{\text{MV}}$). The supervised group includes supervised XGB ($\text{S}_{\text{XGB}}$), supervised RF ($\text{S}_{\text{RF}}$), supervised SVM ($\text{S}_{\text{SVM}}$), a supervised ensemble with a majority voting strategy ($\text{SE}_{\text{MV}}$), and a supervised ensemble using our proposed entropy-based weighted voting strategy ($\text{SE}_{\text{WV}}$). It should be noted that all ensemble models utilize XGB, RF, and SVM as base learners. To ensure consistency, all models are evaluated using the same feature set, comprising contextual embeddings from the fine-tuned ModernBERT model and the set of manually engineered features, including Layout Features, Pattern-Specific Item Features, and Metadata Features.
	%\begin{table*}[!t]
	%	\centering
	%	\caption{Performance comparison of SSE with other classification models}
	%	\label{tab:6}
	%	\resizebox{\linewidth}{!}{%
		%		\begin{tabular}{|c|ccc|ccc|ccc|ccc|ccc|ccc|ccc|ccc|ccc|ccc|}
			%			\hline
			%			\textbf{\begin{tabular}[c]{@{}c@{}}Feature\\ Extraction\\ Method\end{tabular}} &
			%			\multicolumn{3}{c|}{$\text{S}_{\text{SVM}}$} & 
			%			\multicolumn{3}{c|}{$\text{S}_{\text{RF}}$} & 
			%			\multicolumn{3}{c|}{$\text{S}_{\text{XGB}}$} & 
			%			\multicolumn{3}{c|}{$\text{SE}_{\text{MV}}$} & 
			%			\multicolumn{3}{c|}{$\text{SE}_{\text{WV}}$} & 
			%			\multicolumn{3}{c|}{$\text{SS}_{\text{SVM}}$} &
			%			\multicolumn{3}{c|}{$\text{SS}_{\text{RF}}$} &
			%			\multicolumn{3}{c|}{$\text{SS}_{\text{XGB}}$} &
			%			\multicolumn{3}{c|}{$\text{SSE}_{\text{MV}}$} &
			%			\multicolumn{3}{c|}{$\text{SSE}_{\text{WV}}$} \\ \cline{2-31}
			%			& Acc & F1 & TMCC & 
			%			Acc & F1 & TMCC & 
			%			Acc & F1 & TMCC & 
			%			Acc & F1 & TMCC & 
			%			Acc & F1 & TMCC & 
			%			Acc & F1 & TMCC &
			%			Acc & F1 & TMCC &
			%			Acc & F1 & TMCC &
			%			Acc & F1 & TMCC &
			%			Acc & F1 & TMCC \\ \hline
			%			
			%			\multirow{1}{*}{\begin{tabular}[c]{@{}c@{}}Finetuned\\ ModernBERT\\ \& Manual Features\end{tabular}} 
			%			& 0.85412 & 0.86391 & 0.85402
			%			& 0.86681 & 0.87767 & 0.86594
			%			& 0.87526 & 0.87683 & 0.87591
			%			& 0.88372 & 0.88518 & 0.88402
			%			& 0.89429 & 0.89627 & 0.89425
			%			& 0.90486 & 0.90835 & 0.90483
			%			& 0.92389 & 0.92857 & 0.92356
			%			& 0.93869 & 0.94165 & 0.93857
			%			& 0.94080 & 0.94239 & 0.94137
			%			& \textbf{0.95250} & \textbf{0.95471} & \textbf{0.95247} \\ \hline
			%		\end{tabular}%
		%	}
	%\end{table*}
	\begin{table}[!t]
		\centering
		\caption{Performance comparison of SSE with other classification models}
		\label{tab:6}
		\resizebox{\columnwidth}{!}{%
			\begin{tabular}{|c|c|c|c|c|}
				\hline
				\textbf{\begin{tabular}[c]{@{}c@{}}Feature Extraction\\ Method\end{tabular}} & 
				\textbf{\begin{tabular}[c]{@{}c@{}}Classification\\ Models\end{tabular}} & 
				\textbf{Accuracy} & 
				\textbf{F1 Score} & 
				\textbf{TMCC} \\ \hline
				
				\multirow{10}{*}{\begin{tabular}[c]{@{}c@{}}Finetuned\\ ModernBERT\\ \& Manual\\ Features\end{tabular}} 
				& $\text{S}_{\text{SVM}}$      & 0.85412 & 0.86391 & 0.85402 \\
				& $\text{S}_{\text{RF}}$       & 0.86681 & 0.87767 & 0.86594 \\
				& $\text{S}_{\text{XGB}}$      & 0.87526 & 0.87683 & 0.87591 \\
				& $\text{SE}_{\text{MV}}$      & 0.88372 & 0.88518 & 0.88402 \\
				& $\text{SE}_{\text{WV}}$      & 0.89429 & 0.89627 & 0.89425 \\
				& $\text{SS}_{\text{SVM}}$     & 0.90486 & 0.90835 & 0.90483 \\
				& $\text{SS}_{\text{RF}}$      & 0.92389 & 0.92857 & 0.92356 \\
				& $\text{SS}_{\text{XGB}}$     & 0.93869 & 0.94165 & 0.93857 \\
				& $\text{SSE}_{\text{MV}}$     & 0.94080 & 0.94239 & 0.94137 \\
				& $\text{SSE}_{\text{WV}}$     & \textbf{0.95250} & \textbf{0.95471} & \textbf{0.95247} \\ \hline
			\end{tabular}%
		}
	\end{table}
    
    Building on the comparative setup described above, Table~\ref{tab:6} shows that the proposed $\text{SSE}_{\text{WV}}$ model significantly outperforms all supervised and semi-supervised baselines across all evaluation metrics. For instance, $\text{SSE}_{\text{WV}}$ achieves an Accuracy of 0.9525, surpassing the semi-supervised ensemble with majority voting ($\text{SSE}_{\text{MV}}$, 0.94080) by 1.24\% and the best-performing supervised ensemble using the same weighting strategy ($\text{SSE}_{\text{WV}}$, 0.89429) by 6.51\%. The accuracy gains are more substantial when compared to individual supervised models such as $\text{S}_{\text{XGB}}$ (0.87526), $\text{S}_{\text{RF}}$ (0.86681) and $\text{S}_{\text{SVM}}$ (0.85412), with improvements ranging from 8.82\% to 11.52\%. In terms of F1 Score, $\text{SSE}_{\text{WV}}$ attains 0.95471, outperforming $\text{SSE}_{\text{MV}}$ (0.94239) and $\text{SE}_{\text{MV}}$ (0.89627) by 1.31\% and 6.52\%, respectively. It also surpasses semi-supervised single learners, such as $\text{SS}_{\text{XGB}}$ (0.94165), $\text{SS}_{\text{RF}}$ (0.92857), and $\text{SS}_{\text{SVM}}$ (0.90835) by margins of 1.39\%, 2.82\%, and 5.10\%, respectively. The same trend is observed for TMCC, where $\text{SSE}_{\text{WV}}$ leads with a score of 0.95247, outperforming $\text{SSE}_{\text{MV}}$ (0.94137), $\text{SE}_{\text{MV}}$ (0.89425), and all individual classifiers, such as $\text{S}_{\text{XGB}}$ (0.87591), $\text{S}_{\text{RF}}$ (0.86594), $\text{S}_{\text{SVM}}$ (0.85402), $\text{SS}_{\text{XGB}}$ (0.93857), $\text{SS}_{\text{RF}}$ (0.92356), and $\text{SS}_{\text{SVM}}$ (0.90483). These improvements highlight the dual advantage of $\text{SSE}_{\text{WV}}$, including the proposed entropy-based weighting, which prioritizes more confident BLs and its ability to exploit unlabeled data through self-training. Together, these mechanisms contribute to the proposed $\text{SSE}_{\text{WV}}$'s superior performance when classifying diverse and noisy illicit content.
	
	Continuing our evaluation of the proposed framework, we focus on its second stage, which classifies sales-related documents into three distinct categories, including drug sales, weapon sales, and credential sales. This stage consists of three dedicated semi-supervised XGB ($\text{SS}_{\text{XGB}}$) models, each tailored to one of the target categories. To assess their effectiveness, each classifier is compared against a set of baseline models, including supervised SVM ($\text{S}_{\text{SVM}}$), supervised RF ($\text{S}_{\text{RF}}$), supervised XGB ($\text{S}_{\text{XGB}}$), semi-supervised SVM ($\text{SS}_{\text{SVM}}$), and semi-supervised RF ($\text{SS}_{\text{RF}}$). As in the first stage, all models utilize the same feature set, combining contextual embeddings from the fine-tuned ModernBERT model with manually engineered features. The evaluation results, presented in Table~\ref{tab:7}, report the Accuracy, F1 Score, and TMCC for each model, highlighting their performance in classifying sales content into the mentioned subcategories.
	\begin{table}[!t]
		\centering
		\caption{Performance comparison across sale types using different classification models}
		\label{tab:7}
		\resizebox{\columnwidth}{!}{%
			\begin{tabular}{|c|c|c|c|c|c|}
				\hline
				\textbf{\begin{tabular}[c]{@{}c@{}}Feature Extraction\\ Method\end{tabular}} &
				\textbf{\begin{tabular}[c]{@{}c@{}}Sale\\ Type\end{tabular}} &
				\textbf{\begin{tabular}[c]{@{}c@{}}Classification\\ Models\end{tabular}} &
				\textbf{Accuracy} &
				\textbf{F1 Score} &
				\textbf{TMCC} \\ \hline
				
				\multirow{18}{*}{\begin{tabular}[c]{@{}c@{}}Finetuned\\ ModernBERT\\ \& Manual\\ Features\end{tabular}}
				
				& \multirow{6}{*}{Drug}
				& $\text{S}_{\text{SVM}}$     & 0.91121 & 0.79612 & 0.87357 \\
				& & $\text{S}_{\text{RF}}$     & 0.92178 & 0.82126 & 0.89459 \\
				& & $\text{S}_{\text{XGB}}$    & 0.93869 & 0.84656 & 0.90627 \\
				& & $\text{SS}_{\text{SVM}}$   & 0.94926 & 0.88119 & 0.92622 \\
				& & $\text{SS}_{\text{RF}}$    & 0.95983 & 0.91324 & 0.94356 \\
				& & $\text{SS}_{\text{XGB}}$   & \textbf{0.97118} & \textbf{0.93813} & \textbf{0.95985} \\ \cline{2-6}
				
				& \multirow{6}{*}{Weapon}
				& $\text{S}_{\text{SVM}}$     & 0.94922 & 0.83819 & 0.90238 \\
				& & $\text{S}_{\text{RF}}$     & 0.94563 & 0.85393 & 0.91915 \\
				& & $\text{S}_{\text{XGB}}$    & 0.94926 & 0.87097 & 0.92294 \\
				& & $\text{SS}_{\text{SVM}}$   & 0.95983 & 0.90547 & 0.94110 \\
				& & $\text{SS}_{\text{RF}}$    & 0.96967 & 0.92157 & 0.95000 \\
				& & $\text{SS}_{\text{XGB}}$   & \textbf{0.97745} & \textbf{0.94138} & \textbf{0.96385} \\ \cline{2-6}
				
				& \multirow{6}{*}{Credential}
				& $\text{S}_{\text{SVM}}$     & 0.92178 & 0.83528 & 0.89168 \\
				& & $\text{S}_{\text{RF}}$     & 0.92600 & 0.83412 & 0.89616 \\
				& & $\text{S}_{\text{XGB}}$    & 0.93623 & 0.84058 & 0.89817 \\
				& & $\text{SS}_{\text{SVM}}$   & 0.93235 & 0.85185 & 0.90515 \\
				& & $\text{SS}_{\text{RF}}$    & 0.94715 & 0.87685 & 0.92231 \\
				& & $\text{SS}_{\text{XGB}}$   & \textbf{0.95842} & \textbf{0.90448} & \textbf{0.93936} \\ \hline
				
			\end{tabular}%
		}
	\end{table}
	
	Based on the comparative setup described above, Table~\ref{tab:7} shows that the $\text{SS}_{\text{XGB}}$ classifiers in the second stage of the framework consistently outperform their supervised and semi-supervised counterparts across all three illicit sale categories.  For drug sales, $\text{SS}_{\text{XGB}}$ achieves the highest Accuracy (0.97118), outperforming $\text{SS}_{\text{RF}}$ by 1.18\% and $\text{S}_{\text{XGB}}$ by 3.46\%. It also leads in F1 Score (0.93813) and TMCC (0.95985), highlighting its strong balance of precision, recall, and prediction reliability. In weapon sale classification, the $\text{SS}_{\text{XGB}}$ model again achieves the best performance, obtaining an Accuracy of 0.97745, which is 1.17\% higher than $\text{SS}_{\text{RF}}$ and 2.97\% above $\text{S}_{\text{XGB}}$. This improvement underscores the $\text{SS}_{\text{XGB}}$ model’s effectiveness in accurately capturing weapon sales content. Moreover, its F1-score reaches 0.94138, surpassing $\text{SS}_{\text{RF}}$ by 2.15\% and $\text{S}_{\text{XGB}}$ by 8.08\%, ensuring more precise identification of weapon-related instances while minimizing both false positives and false negatives. The TMCC score of $\text{SS}_{\text{XGB}}$ (0.96385) further highlights its superiority, providing a 1.46\% enhancement over $\text{SS}_{\text{RF}}$ and a 4.43\% boost over $\text{S}_{\text{XGB}}$. Regarding stolen credential sales, the $\text{SS}_{\text{XGB}}$ maintains its lead with an Accuracy of 0.95842, outperforming $\text{SS}_{\text{RF}}$ by 1.19\% and $\text{S}_{\text{XGB}}$ by 3.03\%. This consistently high Accuracy confirms the model’s robustness across different types of illicit sales categories. The F1 Score of the $\text{SS}_{\text{XGB}}$ model stands at 0.90448, leading by 3.15\% over $\text{SS}_{\text{RF}}$ and 7.60\% over $\text{S}_{\text{XGB}}$, further validating its strong performance in maintaining a balance between capturing true positive credential-related documents and minimizing false alarms. Additionally, the TMCC score for $\text{SS}_{\text{XGB}}$ reaches 0.93936, outperforming $\text{SS}_{\text{RF}}$ by 1.85\% and $\text{S}_{\text{XGB}}$ by 4.59\%, reflecting a stronger correlation between predicted and actual labels and confirms the model’s higher reliability in classifying credential sale documents.  These results demonstrate the consistent strength of the $\text{SS}_{\text{XGB}}$ classifiers in the second stage, underscoring their effectiveness in handling complex classification tasks across diverse types of illicit marketplace content.
	
	Generally, these results demonstrate that the proposed sequential classification model consistently outperforms various supervised and semi-supervised models in accurately identifying and categorizing various types of illicit marketplace activities. This superior performance highlights the key strengths of our approach, most notably, its ability to utilize large volumes of unlabeled data alongside limited labeled samples through an effective semi-supervised learning strategy. Additionally, the integration of the entropy-based weighted voting mechanism in the first stage significantly enhances the model’s capability in detecting sales-related content, contributing to its overall robustness and reliability.
	
	\subsection{Comparison of the Proposed Model with Prominent Deep Models Across Various Datasets}
	In this experiment, we extend our evaluation by comparing the performance of the proposed sequential classification model against several prominent deep LMs, including $\text{BERT}_{\text{base}}$, $\text{ALBERT}_{\text{base}}$, DarkBERT, BigBird-4096, Longformer-4096, and ModernBERT, across multiple datasets. Among these, DarkBERT is uniquely pretrained from scratch on domain-specific dark web corpora, whereas the remaining models are based on general-purpose pretrained transformers not originally trained on illicit or dark web content. As detailed in previous sections, our proposed model combines manually engineered features with embeddings from ModernBERT, which is fine-tuned on domain-specific data from various illicit platforms. The model is further enhanced by a hierarchical semi-supervised classification strategy, incorporating ensemble learning with an entropy-based weighted voting mechanism. In this comparative study, each deep model is employed as a standalone multi-label classifier tasked with detecting and categorizing illicit marketplace content. To ensure a fair comparison, all models are evaluated using macro-averaged Accuracy, F1 Score, and TMCC. To assess generalization performance, evaluations are conducted across three distinct test sets, including our own test set, along with two well-known benchmark datasets, DUTA and CoDA. Notably, all three datasets are used exclusively for testing, allowing us to evaluate the generalization ability of our proposed model compared to other models across different sources and distributions. To ensure that DUTA and CoDA align with the specific objectives of this study, we applied preprocessing steps that included filtering for English-language documents and relabeling samples according to the four target categories (sale, drug, weapon, and credential). This relabeling process ensures that the evaluation task remains consistent across all datasets and models, maintaining a direct focus on the detection and categorization of illicit marketplace activities. Table~\ref{tab:8} summarizes the composition of the modified DUTA and CoDA datasets after relabeling.  Also, Table~\ref{tab:9} reports the average Accuracy, F1-score, and TMCC for each model on each dataset to provide a comprehensive overview of overall performance.
	\begin{table}[!t]
		\centering
		\caption{Summary Statistics of Modified DUTA and CoDA Datasets Used as Test Sets}
		\label{tab:8}
		\resizebox{\columnwidth}{!}{%
			\begin{tabular}{|l|c|c|c|c|c|}
				\hline
				\textbf{Dataset} & \textbf{Sale} & \textbf{Drug Sale} & \textbf{Weapon Sale} & \textbf{Credential Sale} & \textbf{Total Samples} \\
				\hline
				DUTA & 979 & 317 & 63 & 495 & 6524 \\
				\hline
				CoDA & 2099 & 967 & 597 & 535 & 8855 \\
				\hline
			\end{tabular}
		}
	\end{table}
	
	As shown in Table~\ref{tab:9}, across all three datasets (DUTA, CoDA, and our multi-source test set), the proposed model consistently achieves the highest performance in terms of Accuracy, F1-score, and TMCC. This consistent superiority highlights not only the robustness of the model across different types of data distributions but also the effectiveness of its task-specific design that integrates domain adaptation, hierarchical classification, and semi-supervised ensemble learning.
    Specifically, on the DUTA dataset, our proposed model achieves an F1-score of 0.86568 and TMCC of 0.92058. In comparison, ModernBERT, despite having an input capacity of 8192 tokens, reaches only 0.74293 F1-score and 0.83938 TMCC. This gap of over 12\% in F1-score and more than 8\% in TMCC illustrates the added value of domain-specific fine-tuning, as well as the integration of structural features and task-specific design in the proposed model. Although $\text{BERT}_{\text{base}}$ provides reasonable contextual embeddings with an F1-score of 0.75028 and TMCC of 0.84371, it is limited by its 512-token input length and lack of domain adaptation. This limitation prevents it from capturing the dispersed and complex semantics found in longer illicit content documents. Similarly, Longformer-4096, which supports longer documents using sliding window attention, performs only slightly better than $\text{BERT}_{\text{base}}$, with an F1-score of 0.76528 and TMCC of 0.85436. However, without hierarchical modeling or adaptation to the illicit content domain, it struggles to identify complex patterns. $\text{ALBERT}_{\text{base}}$ improves over both $\text{BERT}_{\text{base}}$ and Longformer with an F1-score of 0.78373 and TMCC of 0.86751, aided by its parameter sharing and regularization. Still, its lightweight design limits its capacity to capture deeper contextual semantics. BigBird-4096 performs reasonably well with 0.80951 F1-score and 0.88436 TMCC, benefiting from sparse attention and scalability. Nevertheless, the absence of domain-specific learning and structured classification prevents it from matching the proposed model’s performance. DarkBERT, which is the only model pretrained on dark web data, achieves stronger results than all general-purpose models, reaching 0.82547 F1-score and 0.89533 TMCC. However, it still falls short by about 4.9\% in F1-score and 2.8\% in TMCC compared to the proposed approach, which benefits from additional semi-supervised learning and ensemble-based decision mechanisms.
    
    The same pattern holds on the CoDA dataset, which includes a larger and more diverse set of illicit content drawn from multiple sources. The proposed model achieves an F1-score of 0.94439 and a TMCC of 0.95880, outperforming ModernBERT by margins of 7.47\% in F1-score and 5.32\% in TMCC. This performance gap again reflects the necessity of domain adaptation, as well as the benefit of incorporating structural features, hierarchical classification, and semi-supervised learning strategies that extend beyond raw model capacity. $\text{BERT}_{\text{base}}$ and $\text{ALBERT}_{\text{base}}$ continue to fall behind, with F1-scores of 0.86578 and 0.89522 and TMCC scores of 0.90089 and 0.92319, which highlights the limitations of using general-purpose models that are not tailored to the linguistic and structural nuances of illicit domain content. Longformer-4096 and BigBird-4096 offer moderately better performance, with F1-scores of 0.88919 and 0.90619 and TMCC scores of 0.91841 and 0.93109, aided by their ability to handle longer sequences. However, they still underperform in F1-score by 6.21\% and 4.22\%, respectively, largely due to their lack of hierarchical supervision, domain-specific refinement, and mechanisms to make use of unlabeled data. Although DarkBERT performs well on CoDA, achieving an F1-score of 0.93043 and a TMCC of 0.94854, the proposed model still surpasses it by 1.5\% in F1-score and 1.03\% in TMCC. This suggests that even with domain-specific pretraining, additional gains in generalization and robustness can be achieved by using a hierarchical classification design, exploiting unlabeled data through self-training, and incorporating ensemble-based decision strategies.
         
	On our multi-source test set, which contains noisier and more heterogeneous content collected from platforms such as Telegram, Pastebin, Reddit, and the deep and dark web, the proposed model again achieves the highest performance with an F1 score of 0.93467 and a TMCC of 0.95388. In contrast, ModernBERT records an F1 score of 0.80930 and a TMCC of 0.86722, suggesting that the ability to process long sequences alone is not sufficient for handling the complexity and irregularity of illicit content across diverse platforms. The observed performance gap of 15.37\% in F1 score and 8.67\% in TMCC clearly illustrates the advantage of integrating domain knowledge with hierarchical modeling and semi-supervised learning strategies. Other baselines such as $\text{BERT}_{\text{base}}$, $\text{ALBERT}_{\text{base}}$, and Longformer-4096 also struggle in this setting, with F1 scores of 0.80820, 0.83016, and 0.83444 and TMCC scores of 0.87268, 0.88592, and 0.88175, respectively. These models are limited when documents are long, contextually fragmented, or semantically ambiguous, as is often the case with underground content. BigBird-4096 achieves a slightly higher F1 score of 0.83933 and a TMCC of 0.89438, benefiting from sparse attention and longer input capacity, but still falls short due to the absence of task-specific adaptation and structured classification. Among the baselines, DarkBERT performs the best with an F1 score of 0.87044 and a TMCC of 0.91242. However, the proposed model exceeds it by 7.38\% in F1 score and 4.54\% in TMCC, reinforcing the point that even domain-specific pretraining must be complemented by layered classification and the ability to exploit unlabeled data in order to deliver robust and generalizable performance across varied and noisy sources.

	%\begin{table*}[!t]
	%	\centering
	%	\caption{Comparison of proposed model with deep models across DUTA, CoDA, and our dataset}
	%	\label{tab:9}
	%	\resizebox{\linewidth}{!}{%
		%		\begin{tabular}{|c|ccc|ccc|ccc|ccc|ccc|ccc|ccc|}
			%			\hline
			%			\textbf{Datasets} & 
			%			\multicolumn{3}{c|}{\textbf{ModernBERT}} & 
			%			\multicolumn{3}{c|}{\textbf{BERT}} & 
			%			\multicolumn{3}{c|}{\textbf{Longformer}} & 
			%			\multicolumn{3}{c|}{\textbf{ALBERT}} & 
			%			\multicolumn{3}{c|}{\textbf{BigBird}} & 
			%			\multicolumn{3}{c|}{\textbf{DarkBERT}} & 
			%			\multicolumn{3}{c|}{\textbf{Proposed}} \\ \cline{2-22}
			%			& Acc & F1 & TMCC & Acc & F1 & TMCC & Acc & F1 & TMCC & Acc & F1 & TMCC 
			%			& Acc & F1 & TMCC & Acc & F1 & TMCC & Acc & F1 & TMCC \\ \hline
			%			
			%			\textbf{DUTA} 
			%			& 0.897 & 0.742 & 0.839 & 0.901 & 0.750 & 0.843 & 0.909 & 0.765 & 0.854 
			%			& 0.920 & 0.783 & 0.867 & 0.932 & 0.809 & 0.884 & 0.940 & 0.825 & 0.895 & 0.957 & 0.865 & 0.920 \\ \hline
			%			
			%			\textbf{CoDA} 
			%			& 0.922 & 0.878 & 0.910 & 0.914 & 0.865 & 0.900 & 0.929 & 0.891 & 0.918 
			%			& 0.932 & 0.895 & 0.918 & 0.939 & 0.906 & 0.931 & 0.955 & 0.930 & 0.948 & 0.964 & 0.944 & 0.958 \\ \hline
			%			
			%			\textbf{Our Dataset} 
			%			& 0.897 & 0.809 & 0.867 & 0.904 & 0.808 & 0.872 & 0.908 & 0.834 & 0.881 
			%			& 0.915 & 0.830 & 0.885 & 0.923 & 0.839 & 0.894 & 0.934 & 0.870 & 0.912 & 0.964 & 0.934 & 0.953 \\ \hline
			%		\end{tabular}%
		%	}
	%\end{table*}
	\begin{table}[!t]
		\centering
		\caption{Comparison of proposed model with deep models across DUTA, CoDA, and Our Dataset}
		\label{tab:9}
		\resizebox{\columnwidth}{!}{%
			\begin{tabular}{|c|ccc|ccc|ccc|}
				\hline
				\textbf{Model} &
				\multicolumn{3}{c|}{\textbf{DUTA}} &
				\multicolumn{3}{c|}{\textbf{CoDA}} &
				\multicolumn{3}{c|}{\textbf{Our Dataset}} \\ \cline{2-10}
				& Acc & F1 & TMCC & Acc & F1 & TMCC & Acc & F1 & TMCC \\ \hline
				
				ModernBERT  & 0.897 & 0.742 & 0.839 & 0.922 & 0.878 & 0.910 & 0.897 & 0.809 & 0.867 \\
				BERT       & 0.901 & 0.750 & 0.843 & 0.914 & 0.865 & 0.900 & 0.904 & 0.808 & 0.872 \\
				Longformer & 0.909 & 0.765 & 0.854 & 0.929 & 0.891 & 0.918 & 0.908 & 0.834 & 0.881 \\
				ALBERT     & 0.920 & 0.783 & 0.867 & 0.932 & 0.895 & 0.918 & 0.915 & 0.830 & 0.885 \\
				BigBird    & 0.932 & 0.809 & 0.884 & 0.939 & 0.906 & 0.931 & 0.923 & 0.839 & 0.894 \\
				DarkBERT   & 0.940 & 0.825 & 0.895 & 0.955 & 0.930 & 0.948 & 0.934 & 0.870 & 0.912 \\
				\textbf{Proposed}   & \textbf{0.957} & \textbf{0.865} & \textbf{0.920} & \textbf{0.964} & \textbf{0.944} & \textbf{0.958} & \textbf{0.964} & \textbf{0.934} & \textbf{0.953} \\ \hline
				
			\end{tabular}%
		}
	\end{table}
	\section{Conclusion}\label{sec:conclusion}
	In this study, we introduced a sequential document classification model designed to detect and categorize marketplace content across various platforms, including the deep web, dark web, Telegram, Reddit, and Pastebin. The first stage incorporates an SSE model with a novel weighted voting strategy to identify sales-related documents, effectively using both labeled and unlabeled data to address the lack of annotated samples. In the second stage, three semi-supervised XGB models classify the detected sales documents into drug, weapon, and credential sales. To capture the linguistic intricacies of such content, we combined fine-tuned ModernBERT embeddings with manually extracted structural features. Fine-tuning ModernBERT on our domain-specific dataset allowed the model to process long documents without truncation, preserving the context needed to detect subtle marketplace cues. This, along with the added structural features, enhanced the model’s ability to handle diverse and complex illicit marketplace content
	In the experimental evaluation, we demonstrated the proposed classification model’s robustness across varying amounts of labeled data, showing consistent improvements in accuracy, F1-score, and TMCC as more labeled data was introduced. This confirmed the model’s ability to effectively utilize unlabeled data within a semi-supervised framework. We then evaluated the effectiveness of each stage of the model separately. In the first stage, we conducted a series of experiments comparing the SSE model to various supervised and semi-supervised baselines using different text representation techniques. These results demonstrated the SSE model’s superior performance in detecting general sales-related documents, highlighting its effective use of unlabeled data and resilience with limited supervision. In the second stage, we assessed the three semi-supervised XGB models for classifying sales-related documents into drug, weapon, and credential categories. Compared to other supervised and semi-supervised classifiers, these models consistently outperformed the alternatives, underscoring the effectiveness of the hierarchical design. Finally, we benchmarked the complete model pipeline against prominent transformer-based baselines, including BERT, ModernBERT, ALBERT, Longformer, BigBird, and DarkBERT, on the DUTA, CoDA, and our multi-source test datasets. The proposed model achieved higher average accuracy, F1-score, and TMCC across all datasets, demonstrating its robustness and adaptability
	
	The results show that the proposed model offers an effective solution for detecting and categorizing illicit marketplace content, successfully addressing challenges related to limited labeled data, long-document processing, and the noisy and dynamic nature of such material. This two-stage sequential classification framework demonstrates strong adaptability and accuracy across diverse datasets, making it well-suited for handling the complex and often concealed language used across underground digital platforms. Future work could focus on expanding the range of monitored categories, refining domain-specific components, and integrating risk assessment modules to enhance the model’s ability to identify emerging threats and support proactive monitoring of evolving activities in this domain.
	
	\section*{Acknowledgments}
	We extend our gratitude to the Digital Finance Cooperative Research Centre (DFCRC) and Cyber Intelligence House (CIH) Company for their invaluable contributions and assistance. Their dedicated efforts have been instrumental in the advancement of our research.


       \bibliographystyle{IEEEtran}
       \bibliography{IEEE}

	%
	%\begin{thebibliography}{1}
	%\bibliographystyle{IEEEtran}
	%
	%\bibitem{ref1}
	%{\it{Mathematics Into Type}}. American Mathematical Society. [Online]. Available: https://www.ams.org/arc/styleguide/mit-2.pdf
	%
	%\bibitem{ref2}
	%T. W. Chaundy, P. R. Barrett and C. Batey, {\it{The Printing of Mathematics}}. London, U.K., Oxford Univ. Press, 1954.
	%
	%\bibitem{ref3}
	%F. Mittelbach and M. Goossens, {\it{The \LaTeX Companion}}, 2nd ed. Boston, MA, USA: Pearson, 2004.
	%
	%\bibitem{ref4}
	%G. Gr\"atzer, {\it{More Math Into LaTeX}}, New York, NY, USA: Springer, 2007.
	%
	%\bibitem{ref5}M. Letourneau and J. W. Sharp, {\it{AMS-StyleGuide-online.pdf,}} American Mathematical Society, Providence, RI, USA, [Online]. Available: http://www.ams.org/arc/styleguide/index.html
	%
	%\bibitem{ref6}
	%H. Sira-Ramirez, ``On the sliding mode control of nonlinear systems,'' \textit{Syst. Control Lett.}, vol. 19, pp. 303--312, 1992.
	%
	%\bibitem{ref7}
	%A. Levant, ``Exact differentiation of signals with unbounded higher derivatives,''  in \textit{Proc. 45th IEEE Conf. Decis.
		%Control}, San Diego, CA, USA, 2006, pp. 5585--5590. DOI: 10.1109/CDC.2006.377165.
	%
	%\bibitem{ref8}
	%M. Fliess, C. Join, and H. Sira-Ramirez, ``Non-linear estimation is easy,'' \textit{Int. J. Model., Ident. Control}, vol. 4, no. 1, pp. 12--27, 2008.
	%
	%\bibitem{ref9}
	%R. Ortega, A. Astolfi, G. Bastin, and H. Rodriguez, ``Stabilization of food-chain systems using a port-controlled Hamiltonian description,'' in \textit{Proc. Amer. Control Conf.}, Chicago, IL, USA,
	%2000, pp. 2245--2249.
	%
	%\end{thebibliography}

\end{document}